\documentclass[10pt]{article} 
\usepackage[preprint]{tmlr}

\usepackage{amsmath,amsfonts,bm}

\def\eqref#1{equation~\ref{#1}}

\def\1{\bm{1}}

\DeclareMathAlphabet{\mathsfit}{\encodingdefault}{\sfdefault}{m}{sl}
\SetMathAlphabet{\mathsfit}{bold}{\encodingdefault}{\sfdefault}{bx}{n}

\usepackage{hyperref}
\usepackage{url}

\usepackage{algorithm}
\usepackage{algorithmic}
\usepackage{enumitem}
\usepackage{microtype}
\usepackage{booktabs}
\usepackage{graphicx}
\usepackage{amsmath}
\usepackage{amsthm}
\usepackage{subcaption}
\usepackage[table]{xcolor} 
\usepackage[many]{tcolorbox}
\usepackage{wrapfig}

\usepackage{array}
\usepackage{tabularx}
\usepackage{colortbl}
\newcolumntype{C}{>{\centering\arraybackslash}X}

\newtheorem{proposition}{Proposition}
\definecolor{rowhl}{gray}{0.92}

\newcommand{\grouphead}[2]{%
  \multicolumn{#1}{@{}l@{}}{\textit{#2}}%
}

\newcommand{\meanstd}[2]{#1\,{\scriptsize$\pm$\,#2}}
\newcommand{\method}{FPA}

\title{Future Policy Approximation for \\ Offline Reinforcement Learning in LLM Reasoning}

\author{\name Minjae Oh, Yunho Choi, Dongmin Choi, Yohan Jo$^{\dagger}$ \\
  \addr Graduate School of Data Science, Seoul National University \\ {\tt kosair@snu.ac.kr, dbsgh7177@snu.ac.kr, chrisandjj@naver.com, yohan.jo@snu.ac.kr}\\
  $^{\dagger}$Corresponding author.
}

\def\month{MM}  
\def\year{YYYY} 
\def\openreview{\url{https://openreview.net/forum?id=XXXX}} 

\begin{document}

\maketitle

\begin{abstract}
Reinforcement learning (RL) has emerged as a key driver of post-training for complex reasoning in large language models (LLMs), yet online RL introduces substantial instability and computational overhead. Offline RL offers a compelling alternative by decoupling generation from training; however, offline algorithms for reasoning remain under-optimized relative to their online counterparts. We revisit the potential of policy-gradient-style offline RL and address a central challenge in offline learning: \emph{gradient entanglement}. In long-horizon reasoning trajectories, correct and incorrect solutions share substantial token overlap, causing gradient updates from incorrect trajectories to suppress tokens that are also critical for correct ones. We propose \textbf{Future Policy Approximation (\method{})}, a simple offline policy-gradient method that weights gradients using an estimate of the future policy rather than the current policy, enabling \emph{proactive} gradient reweighting. We estimate the future policy through logit-space extrapolation. Across three models, seven mathematical reasoning benchmarks, and three code-generation benchmarks, \method{} consistently improves over strong offline baselines, including DPO, RPO, KTO, and vanilla offline RL. \method{} stabilizes long-horizon training, where vanilla objectives degrade, and achieves accuracy comparable to state-of-the-art RLVR methods such as GRPO and DAPO at a fraction of the GPU hours.
\footnote{We will release our code publicly upon publication.}
\end{abstract}

\section{Introduction}
Driven by recent advances in reinforcement learning (RL) for post-training, large language models (LLMs) have demonstrated remarkable capabilities in complex, multi-step reasoning, with mathematics and coding as representative domains \citep{jaech2024openai, yang2025qwen3}. Reinforcement Learning from Verifiable Rewards (RLVR) \citep{lambert2024tulu, guo2025deepseek} has become the dominant approach for eliciting Chain-of-Thought (CoT) reasoning \citep{cot2022} in LLMs, most notably through algorithms such as Group Relative Policy Optimization (GRPO) \citep{shao2024deepseekmath} and Decoupled Clip and Dynamic Sampling Policy Optimization (DAPO) \citep{yu2025dapo}. Despite this success, online RL remains notoriously unstable, frequently suffering from reward collapse or divergence, and computationally expensive, as simultaneous inference and training create a substantial bottleneck \citep{liu2025rlcollapse, zhang2026mismatch, yao2025rolloutmismatch}. These challenges motivate offline RL, which fully decouples inference from training: a fixed dataset is generated once and reused across multiple training runs, providing greater stability, computational efficiency, and engineering simplicity. Recent work has begun to demonstrate the viability of offline algorithms as alternatives to online RLVR in reasoning domains \citep{wang2025offline, lu2026pcl}, yet offline RL for reasoning remains less studied and less optimized than its online counterpart, leaving substantial room for improved offline objectives.

We first revisit a simple REINFORCE-style policy-gradient objective \citep{williams1992simple} for offline reasoning (see \S~\ref{sec:prelim_Off-RL}), following recent work \citep{lu2026pcl}, and find that, despite its simplicity, it performs competitively with established offline learning algorithms. Building on this strong baseline, we address a central challenge in learning from reasoning trajectories: \emph{gradient entanglement} \citep{yuan2025common}. In long-horizon reasoning trajectories, correct and incorrect solutions inevitably share substantial token overlap (e.g., the same intermediate steps, symbols, and partial derivations), causing gradient updates from incorrect trajectories to inadvertently suppress tokens that are also critical for correct ones. This interference limits performance gains and can eventually lead to model degradation or collapse under extended training (see \S~\ref{sec:prelim_grad}). This problem has been observed when applying Direct Preference Optimization (DPO) \citep{rafailov2024dpo} to mathematical reasoning \citep{lai2024step, pal2024smaug, pang2024iterativerpo}, and related instabilities have recently been reported in online RLVR as well \citep{deng2025negative}.

\begin{figure}[!t]
    \centering
    \includegraphics[width=0.9\textwidth]{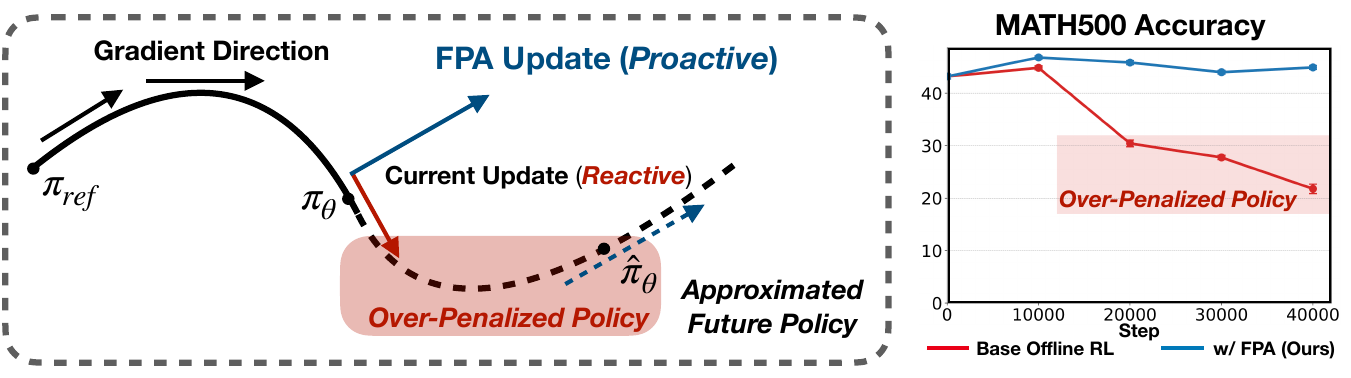}
    \vspace{-0.5em}
    \caption{\textbf{Overview of \method{}.} \textbf{(Left)} Conceptual overview of policy training. Naive updates are reactive, reducing the effect of over-penalization only after it has occurred. \method{} proactively estimates the future policy to anticipate and prevent over-penalization. \textbf{(Right)} On Llama-3.2-3B-Instruct, naive offline RL training on mathematical reasoning leads to performance collapse due to over-penalization, as evidenced by an abrupt drop in accuracy, whereas \method{} maintains stable training throughout.}
    \label{fig:main_fig}
    \vspace{-1em}
\end{figure}

We propose \textbf{Future Policy Approximation (\method{})}, a simple method for improving offline RL for reasoning (see \S~\ref{sec:method}). Existing offline policy-gradient objectives modulate gradient magnitudes using the current policy: as the probability of an incorrect sample decreases, its gradient contribution is down-weighted to prevent over-penalization, while higher-probability correct samples receive stronger reinforcing updates. However, this mechanism is inherently \emph{reactive}: damping occurs only after the probability of an incorrect trajectory has already decreased, by which point shared tokens critical to correct trajectories may already have been over-penalized. \method{} instead takes a \emph{proactive} approach by weighting gradients using an estimate of the future policy (see Figure~\ref{fig:main_fig} (left)). To construct this estimate, we build on observations that RL post-training induces approximately linear trajectories in logit space \citep{liu2024decoding} and extrapolate the current policy along the direction of movement from the reference policy. We directly validate this approximation by comparing the extrapolated policy with policies obtained at later training checkpoints, finding that it more closely matches the actual future policy than the current policy does across multiple future checkpoints. Our ablations further show that the gains arise primarily from proactively suppressing harmful updates from incorrect trajectories, consistent with the proposed gradient-entanglement mechanism (see \S~\ref{sec:exp_further}).

We evaluate \method{} across three models from the Qwen and Llama families and seven mathematical reasoning benchmarks, including the commonly used GSM8K~\citep{cobbe2021gsm8k}, MATH500~\citep{hendrycks2021math}, and AIME. \method{} consistently improves over strong offline baselines, including Direct Preference Optimization (DPO)~\citep{rafailov2024dpo}, Reasoning Preference Optimization (RPO)~\citep{pang2024iterativerpo}, Kahneman-Tversky Optimization (KTO)~\citep{Ethayarajh2024KTO}, and vanilla offline policy-gradient RL~\citep{lu2026pcl} (see \S~\ref{sec:exp_math}). \method{} stabilizes long-horizon training where vanilla offline objectives degrade, maintaining performance under prolonged optimization (see Figure~\ref{fig:main_fig} (right)). We further show that its benefits persist under step-level preference supervision, indicating that \method{} is not restricted to trajectory-level feedback. Compared with online RLVR such as GRPO and DAPO, \method{} achieves competitive accuracy at a fraction of the total GPU cost (see \S~\ref{sec:exp_further}). Across three code-generation benchmarks, \method{} consistently improves over offline baselines, suggesting that its benefits extend beyond mathematical reasoning (see \S~\ref{sec:exp_coding}). Furthermore, we validate these gains across three random seeds in both mathematical reasoning and code generation, showing that the improvements persist across independent training runs. Overall, \method{} demonstrates that proactive future-policy signals can improve offline RL training dynamics across supervision granularities and reasoning domains while retaining the computational and data-reuse benefits of offline learning.

\section{Preliminaries}
\subsection{Offline RL}
\label{sec:prelim_Off-RL}

\paragraph{Definition.}
We first consider a REINFORCE-style policy-gradient objective \citep{williams1992simple} for offline RL. Given a fixed dataset $\mathcal{D}=\{(x,y,R)\}$ and a policy $\pi_\theta(y\mid x)$, we define
\begin{equation}
    \label{eq:RL}
    \mathcal{J}(\theta)
    =
    \mathbb{E}_{(x,y,R)\sim\mathcal{D}}
    \left[
        R \times
        \pi_\theta(y\mid x)^{\frac{1}{|y|}}
    \right],
\end{equation}
where $R\in\{+1,-1\}$ denotes the reward for correct and incorrect trajectories, respectively, and $|y|$ denotes the number of tokens in $y$. Following recent work on stabilizing long-horizon policy optimization \citep{zheng2025gspo}, we use the length-normalized trajectory probability $\pi_\theta(y\mid x)^{1/|y|}$ rather than the raw sequence probability. We refer to this objective as \textbf{Off-RL}. This formulation has demonstrated competitive empirical performance in recent work \citep{lu2026pcl}. Furthermore, the recently proposed preference-learning algorithm SimPER \citep{xiao2025simper} can be understood as a special case of Off-RL (see \S~\ref{app:simper_Off-RL}) and has been adopted for industry-scale post-training \citep{research2025exaone4,research2025exaonedeep}.

\paragraph{Training Dynamics.}
The gradient of Eq.~\ref{eq:RL} naturally decomposes into three factors: the reward, the model probability, and the score function:
\begin{align}
    \label{eq:RLgrad}
    \nabla_\theta \mathcal{J}(\theta)
    =
    \mathbb{E}_{(x,y,R)\sim\mathcal{D}}
    \left[
        \frac{1}{|y|}
        \underbrace{R}_{\text{reward}}
        \times
        \underbrace{\pi_\theta(y \mid x)^{\frac{1}{|y|}}}_{\text{probability}}
        \times
        \underbrace{\nabla_\theta \log \pi_\theta(y \mid x)}_{\text{score function}}
    \right].
\end{align}
The three factors play distinct roles. The score function $\nabla_\theta \log \pi_\theta(y \mid x)$ determines the direction and magnitude of each update. The reward $R$ determines whether this signal reinforces or suppresses a trajectory: correct trajectories are reinforced, whereas incorrect trajectories are penalized. However, this mechanism alone can lead to instability. Because $|\nabla_\theta \log \pi_\theta(y \mid x)| \propto 1/\pi_\theta(y \mid x)$, the gradient norm for incorrect trajectories can grow as training drives $\pi_\theta(y \mid x)$ downward, creating a vicious cycle of increasingly large gradients from incorrect trajectories and potentially leading to model collapse \citep{mao2024simple}.

The probability term $\pi_\theta(y \mid x)^{1/|y|}$ provides a policy-dependent weighting that naturally mitigates this problem. For incorrect trajectories to which the current policy already assigns low probability, this term down-weights their gradient contributions, preventing excessive penalization. Conversely, for correct trajectories to which the policy assigns high probability, it places greater weight on their updates, encouraging exploitation. Thus, as the policy evolves away from the fixed offline data, the probability term adaptively modulates the contribution of each trajectory according to its likelihood under the current policy.

\subsection{Gradient Entanglement}
\label{sec:prelim_grad}
One major problem in offline learning for reasoning is a phenomenon known as \textit{gradient entanglement} \citep{yuan2025common}. In reasoning domains, correct and incorrect trajectories often share substantial portions of their tokens, such as intermediate calculations, symbols, and reasoning steps. As a result, their score-function gradients, $\nabla_\theta \log \pi_\theta(y\mid x)$, can be highly aligned. However, because learning objectives such as DPO and Off-RL reinforce correct trajectories while suppressing incorrect ones, the induced update directions become opposed. This creates \emph{gradient entanglement}: updates that suppress an incorrect trajectory can simultaneously suppress tokens that are important for correct trajectories.

This issue becomes more severe when the opposing updates have different magnitudes. As discussed in \S~\ref{sec:prelim_Off-RL}, the score-function gradient of an incorrect trajectory can grow as its probability decreases, often resulting in larger gradients for incorrect trajectories than for correct ones. Under gradient entanglement, these larger incorrect gradients can dominate the combined update and overwhelm the reinforcing signal from correct trajectories.

\begin{figure}[!t]
\centering
    \begin{subfigure}[b]{0.45\textwidth}
        \includegraphics[width=\textwidth]{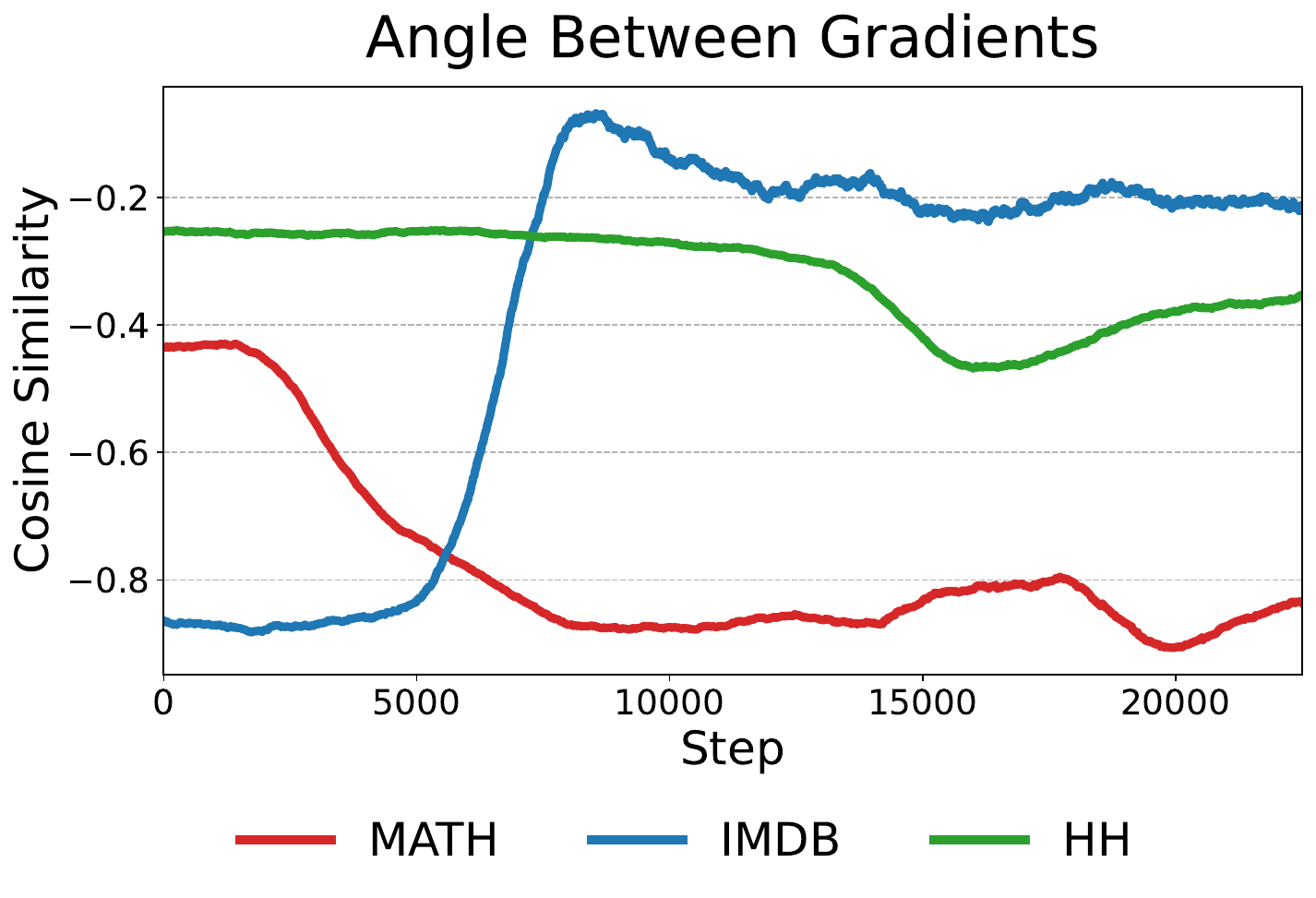}
    \end{subfigure}
    \begin{subfigure}[b]{0.45\textwidth}
        \includegraphics[width=\textwidth]{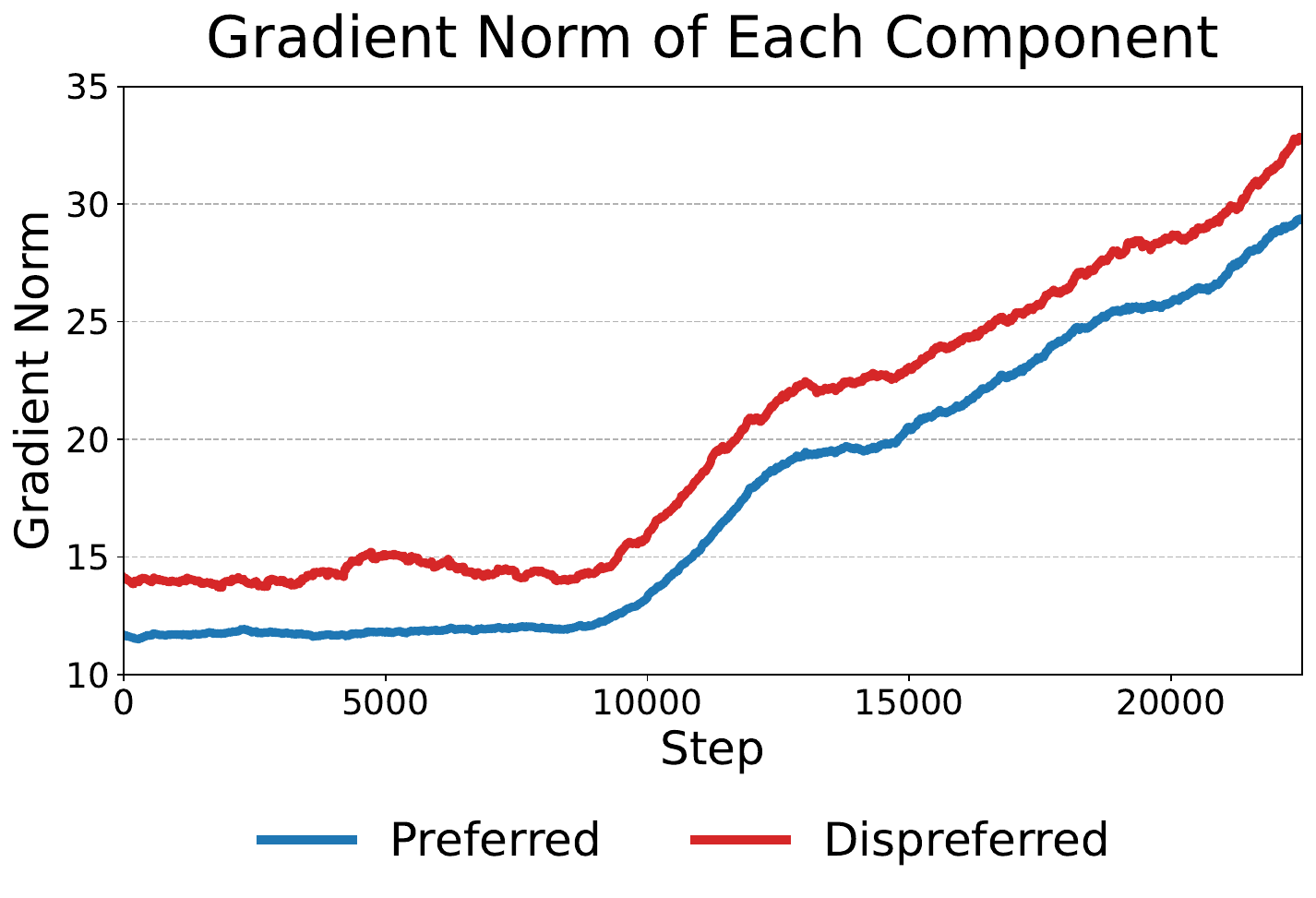}
    \end{subfigure}
    \caption{\textbf{Gradient entanglement across datasets.}
    (\textbf{Left}) The update directions induced by preferred and dispreferred trajectories are substantially more opposed on MATH than on Anthropic Helpful and Harmless (HH) and Stanford IMDb.
    (\textbf{Right}) On MATH, the gradient norms of dispreferred trajectories are consistently larger than those of preferred trajectories.
    Experimental details are provided in \S~\ref{app:pilot_details}.}
    \label{fig:grad_ent}
\end{figure}

We empirically examine this behavior across three datasets with different trajectory characteristics: the mathematical reasoning dataset MATH~\citep{hendrycks2021math} and two shorter-form datasets, Stanford IMDb~\citep{maas2011learning} and Anthropic Helpful and Harmless (HH)~\citep{bai2022training}. As shown in Figure~\ref{fig:grad_ent} (left), the update directions induced by preferred and dispreferred trajectories are substantially more opposed on MATH than on IMDb and HH, indicating stronger gradient entanglement in mathematical reasoning. Moreover, Figure~\ref{fig:grad_ent} (right) shows that the gradient norms of incorrect trajectories on MATH are consistently larger than those of correct trajectories. Together, these results indicate that mathematical reasoning exhibits both strong gradient interference and a systematic magnitude imbalance toward incorrect trajectories, making long-horizon reasoning particularly susceptible to gradient entanglement. We further validate this effect during training, showing in \S~\ref{sec:exp_math} that the probabilities of correct trajectories decrease alongside those of incorrect trajectories under Off-RL. These observations suggest that gradients from incorrect trajectories require stronger and more systematic regularization than that provided by the current probability weighting, particularly before they grow large enough to dominate the combined update under gradient entanglement.

\section{Future Policy Approximation for Offline Reinforcement Learning} \label{sec:method} 
\subsection{Intuition and Definition} 
\label{sec:method_fpa_def} 
We introduce \textbf{Future Policy Approximation (\method{})} to address gradient entanglement in offline RL for reasoning. From Eq.~\ref{eq:RLgrad}, the probability term modulates gradient magnitudes based on the current policy $\pi_\theta$. However, this dynamic is inherently \emph{reactive}: damping occurs only after the policy has already assigned low probability to a trajectory, by which point useful tokens shared under gradient entanglement may already have been over-penalized. To remedy this, we make the weighting \emph{proactive} by replacing the current-policy probability with a future-policy estimate $\hat{\pi}_{\theta,\lambda}$:
\begin{equation} 
\label{eq:fpa_grad} 
\nabla_\theta \mathcal{J}_{\mathrm{FPA}}(\theta) = \mathbb{E}_{(x,y,R)\sim\mathcal{D}} \left[ \frac{1}{|y|}R \times \operatorname{sg}\!\left[ \hat{\pi}_{\theta,\lambda}(y\mid x)^{\frac{1}{|y|}} \right] \times \nabla_\theta \log \pi_\theta(y\mid x) \right], 
\end{equation}
where $\operatorname{sg}[\cdot]$ denotes the stop-gradient operator and $\lambda$ is a hyperparameter. Thus, $\hat{\pi}_{\theta,\lambda}$ determines the magnitude of each update, while the gradient continues to be computed with respect to the current policy $\pi_\theta$. By weighting gradients using $\hat{\pi}_{\theta,\lambda}$, \method{} anticipates probability changes before they occur. For example, if an incorrect trajectory currently has relatively high probability but the future policy predicts a substantial decrease along the observed training direction, Off-RL down-weights its gradient only after the probability has decreased. \method{} instead regularizes the gradient earlier, functioning as an \emph{early brake} before shared useful tokens are excessively penalized. Conversely, when the probability of a correct trajectory is increasing along the observed training direction, $\hat{\pi}_{\theta,\lambda}$ assigns it a larger weight and can strengthen its reinforcing update. In practice, however, our ablations show that \method{}'s primary benefit comes from suppressing harmful gradients from incorrect trajectories (see \S~\ref{sec:exp_further}).

\subsection{Approximating the Future Policy} 
\label{sec:fpa2} 
We follow recent work showing that LLM post-training, including RL and DPO, induces approximately linear movement in logit space over the course of training, and that further-trained models can be approximated through logit-space extrapolation \citep{liu2024decoding, kim2024spread, huang2026on, wang2026not}. Formally, at timestep $t$, let $h_\theta(x,y_{<t})$ and $h_\text{ref}(x,y_{<t})$ denote the next-token logits of the current policy and the fixed reference policy, respectively. We define the extrapolated future policy with strength $\lambda \geq 0$ as
\begin{align} 
\label{eq:future_policy} 
\hat{\pi}_{\theta,\lambda}(\cdot \mid x,y_{<t}) &= \operatorname{softmax}\!\left( h_\theta(x,y_{<t}) + \lambda\!\left( h_\theta(x,y_{<t}) - h_{\mathrm{ref}}(x,y_{<t}) \right)\right) \nonumber\\ &= \operatorname{softmax}\!\left( (1+\lambda)h_\theta(x,y_{<t}) - \lambda h_\text{ref}(x,y_{<t}) \right). 
\end{align} 
That is, we start from the current logits and extrapolate further in the direction in which training has moved the policy away from the reference policy. When $\lambda=0$, $\hat{\pi}_{\theta,\lambda}=\pi_\theta$, and \method{} reduces exactly to Off-RL; larger values of $\lambda$ extrapolate farther along the observed training direction. The trajectory-level future probability used in Eq.~\ref{eq:fpa_grad} is obtained by factorizing over tokens:
\begin{equation} \hat{\pi}_{\theta,\lambda}(y\mid x) = \prod_{t=1}^{|y|} \hat{\pi}_{\theta,\lambda}(y_t\mid x,y_{<t}). 
\end{equation} 
Equivalently, expressing Eq.~\ref{eq:future_policy} in terms of the current and reference policies yields
\begin{equation} 
\label{eq:future_policy_prob} 
\hat{\pi}_{\theta,\lambda}(\cdot \mid x,y_{<t}) \propto \pi_\theta(\cdot \mid x,y_{<t})^{1+\lambda} \pi_\text{ref}(\cdot \mid x,y_{<t})^{-\lambda}. 
\end{equation} 
The key assumption behind \method{} is that the direction from the reference policy to the current policy contains useful information about subsequent policy movement. We directly validate this assumption in \S~\ref{sec:exp_further} by comparing $\hat{\pi}_{\theta,\lambda}$ with policies obtained at later training checkpoints. In principle, \method{} can also be implemented with only marginal additional computation: $h_\theta$ is already obtained during training, while $h_\text{ref}$ can be precomputed and cached for the fixed offline dataset.

\section{Experiments} 
\label{sec:experiments}

\subsection{Experimental Setup}
\label{sec:exp_setup}
We employ three foundation models: Qwen3-4B-Base~\citep{yang2025qwen3} as a general-purpose base model, Qwen2.5-Math-7B~\citep{yang2024qwenmath} as a math-specialized model, and Llama-3.2-3B-Instruct~\citep{grattafiori2024llama} as an instruction-tuned model. Unless otherwise stated, we use Qwen3-4B-Base as the default model. For mathematical reasoning, we construct the offline training dataset from the MATH training set~\citep{hendrycks2021math}, sampling 8 trajectories per problem at temperature 0.7, following the group size commonly used in RLVR training~\citep{vonwerra2020trl}. We evaluate on seven mathematical reasoning benchmarks: GSM8K~\citep{cobbe2021gsm8k}, MATH500 (a 500-problem subset of the MATH test set)~\citep{hendrycks2021math}, Math-Perturb~\citep{huang2025mathp}, GaoKao23~\citep{liao2024gaokao}, OlympiadBench~\citep{he2024olympiadbench}, AMC23/24, and AIME24/25/26. We report Pass@1 accuracy using 8 independently sampled trajectories per problem at temperature 0.7, except for AMC and AIME, for which we use 32 trajectories per problem. We additionally repeat the main mathematical experiments with three random seeds and report the mean and sample standard deviation across runs to assess robustness. The \method{} extrapolation strength $\lambda$ in Eq.~\ref{eq:future_policy} is set to 2 for Qwen3-4B-Base and 1 for other models based on validation-set performance (see \S~\ref{sec:exp_hyperparameter} and \S~\ref{app:hyperparameter}). \method{} adds minimal computational overhead (see \S~\ref{app:exp_details}).

To evaluate whether \method{} extends beyond mathematical reasoning, we additionally conduct experiments in the code-generation domain. We construct the offline training data from TACO~\citep{li2023taco} and APPS~\citep{hendrycks2021apps}, and evaluate on three established code-generation benchmarks: MBPP+~\citep{liu2023your}, BigCodeBench (BCB)~\citep{zhuo2025bigcodebench}, and LiveCodeBench v6 (LCB)~\citep{jain2025livecodebench}. We use Qwen3-4B-Base and Llama-3.2-3B-Instruct and follow the same evaluation protocol as in the mathematical experiments, sampling $n=8$ trajectories per problem. To assess robustness to training randomness, we repeat the main coding experiments with three random seeds and report the mean and sample standard deviation across runs (see \S~\ref{sec:exp_coding} and \S~\ref{app:multiseed}).

We compare \method{} against a broad set of established offline learning methods, including Supervised Fine-Tuning (SFT)~\citep{yuan2023rft}, which serves as a standard supervised reference; Direct Preference Optimization (DPO)~\citep{rafailov2024dpo}, a widely used offline RL method; Reasoning Preference Optimization (RPO)~\citep{pang2024iterativerpo} and DPO-Positive (DPOP)~\citep{pal2024smaug}, which both improve upon DPO in reasoning domains; Kahneman-Tversky Optimization (KTO)~\citep{Ethayarajh2024KTO}, a strong binary preference-learning algorithm; an offline version of Policy Optimization via Optimal Advantage Regression (A*-PO)~\citep{brantley2025accelerating}, a recently proposed reasoning-focused policy optimization method; and Off-RL~\citep{lu2026pcl}, our policy-gradient RL baseline. Further details on the baselines and experimental settings are provided in \S~\ref{app:exp_details}.

\subsection{Mathematical Experiment Results}
\label{sec:exp_math}
\begin{table*}[!tb]
\centering
\caption{\textbf{Benchmark results.} All results are Pass@1 accuracy. The best performance within each model block is \textbf{bolded}, and the second-best performance is \underline{underlined}.}
\small
\setlength{\tabcolsep}{3pt}
\renewcommand{\arraystretch}{1.10}
\begin{tabularx}{\textwidth}{@{}l *{8}{C}@{}}
\toprule
\textbf{Method}
& \textbf{GSM8K}
& \textbf{MATH500}
& \textbf{MATH-P}
& \textbf{GaoKao}
& \textbf{Olym.}
& \textbf{AMC}
& \textbf{AIME}
& \textbf{Avg.} \\
\midrule
\grouphead{9}{Qwen3-4B-Base} \\
Base
& 75.52 & 56.83 & 33.18 & 43.96 & 25.40 & 29.78 & 6.60 & 38.75 \\
SFT
& 81.30 & 58.75 & 34.57 & 46.72 & 26.35 & 31.18 & 6.67 & 40.79 \\
DPO
& 85.05
& \underline{69.35}
& \underline{43.66}
& \underline{55.13}
& \underline{32.76}
& \underline{39.39}
& \underline{9.58}
& \underline{47.85} \\
RPO
& \underline{88.61} & 64.70 & 38.98 & 52.37 & 30.37 & 33.54 & 8.50 & 45.30 \\
DPOP
& 82.30 & 64.88 & 41.46 & 51.36 & 30.06 & 35.60 & 8.92 & 44.94 \\
KTO
& 86.50 & 67.60 & 40.24 & 53.44 & 31.44 & 36.40 & 7.99 & 46.23 \\
A*-PO
& 86.65 & 63.82 & 42.57 & 48.12 & 31.40 & 33.30 & 7.88 & 44.82 \\
Off-RL
& 86.80 & 66.70 & 41.42 & 52.76 & 30.72 & 36.91 & 8.50 & 46.26 \\
\rowcolor{rowhl}
\hspace{0.5em}\textbf{w/ \method{}}
& \textbf{91.60}
& \textbf{74.50}
& \textbf{48.40}
& \textbf{59.60}
& \textbf{35.40}
& \textbf{42.80}
& \textbf{10.10}
& \textbf{51.77} \\
\midrule
\grouphead{9}{Qwen2.5-Math-7B} \\
Base
& 77.17 & 63.65 & 39.79 & 47.73 & 28.96 & 36.79 & 10.80 & 43.56 \\
SFT
& 80.75 & 66.80 & 40.08 & 50.88 & 30.63 & 40.76 & 10.21 & 45.73 \\
DPO
& \underline{82.25} & 67.90 & 38.06 & 49.97 & 29.33 & 40.01 & 11.56 & 45.58 \\
RPO
& 81.97 & \underline{68.73} & 38.37 & 50.36 & 28.76 & 40.71 & 10.63 & 45.65 \\
DPOP
& 81.58 & 68.50 & 43.23 & 51.85 & 30.87 & 40.30 & \textbf{13.75} & 47.15 \\
KTO
& 82.20 & 68.60 & 38.37 & 52.21 & 29.09 & \textbf{43.09} & 10.97 & 46.36 \\
A*-PO
& 79.75 & 67.80 & 42.30 & 50.29 & \underline{32.17} & 38.85 & 10.90 & 46.01 \\
Off-RL
& 80.39
& \underline{68.73}
& \underline{43.91}
& \underline{53.05}
& 31.72
& 41.99
& 11.84
& \underline{47.38} \\
\rowcolor{rowhl}
\hspace{0.5em}\textbf{w/ \method{}}
& \textbf{83.95}
& \textbf{72.55}
& \textbf{44.58}
& \textbf{56.07}
& \textbf{33.50}
& \underline{42.32}
& \underline{12.74}
& \textbf{49.39} \\
\midrule
\grouphead{9}{Llama-3.2-3B-Instruct} \\
Base
& 77.52 & 43.20 & 20.88 & 32.99 & 13.80 & \underline{16.34} & \underline{2.71} & 29.63 \\
SFT
& 76.63 & 42.62 & 20.57 & 31.37 & 13.13 & 16.19 & 2.22 & 28.96 \\
DPO
& 78.27 & \underline{44.22} & 21.28 & 32.53 & \underline{13.96} & 15.56 & 1.81 & 29.66 \\
RPO
& 78.24 & 44.20 & \underline{21.42} & 32.95 & 13.50 & 14.86 & 1.25 & 29.49 \\
DPOP
& 71.95 & 33.32 & 19.76 & 27.27 & 7.93 & 12.29 & \textbf{2.74} & 25.04 \\
KTO
& 77.03 & 43.70 & 21.24 & 32.60 & \underline{13.96} & 14.78 & 1.95 & 29.32 \\
A*-PO
& 77.22 & 43.82 & \textbf{21.55} & 33.18 & 13.44 & 16.15 & 2.12 & 29.64 \\
Off-RL
& \underline{78.75}
& 43.75
& 21.13
& \underline{33.90}
& 13.72
& 14.14
& 2.29
& \underline{29.67} \\
\rowcolor{rowhl}
\hspace{0.5em}\textbf{w/ \method{}}
& \textbf{80.54}
& \textbf{46.02}
& 20.70
& \textbf{34.16}
& \textbf{14.13}
& \textbf{18.37}
& 2.08
& \textbf{30.86} \\
\bottomrule
\end{tabularx}
\label{tab:main_results}
\end{table*}

Our primary results are summarized in Table~\ref{tab:main_results}. \method{} achieves the best average performance across all three models and the strongest performance on most individual benchmarks. Notably, vanilla Off-RL (Eq.~\ref{eq:RL}) already performs competitively with the other offline baselines, highlighting the effectiveness of policy-gradient objectives for offline reasoning. \method{} provides further substantial gains over this strong baseline. On Qwen3-4B-Base, \method{} improves the average accuracy by 5.51\% over Off-RL, including gains of 7.80\% on MATH500 and 6.84\% on GaoKao. On Qwen2.5-Math-7B, \method{} improves the average by 2.01\%, including a 3.82\% gain on MATH500. Notably, \method{} also improves Llama-3.2-3B-Instruct---an instruction-tuned model that is known to benefit less from further mathematical post-training~\citep{liu2026tricks}---achieving a 1.19\% average improvement over Off-RL, including gains of 2.27\% on MATH500 and 4.23\% on AMC.

We further verify the robustness of these gains through three-seed experiments on Qwen3-4B-Base (see \S~\ref{app:multiseed}), where \method{} retains the strongest average performance across runs. In addition, our full hyperparameter sweeps for all baselines (see \S~\ref{app:baseline_tuning}) show that \method{} continues to outperform the best-performing configuration of every baseline, indicating that the improvements are not attributable to insufficient baseline hyperparameter tuning.

\begin{figure}[!t]
\centering
    \includegraphics[width=\textwidth]{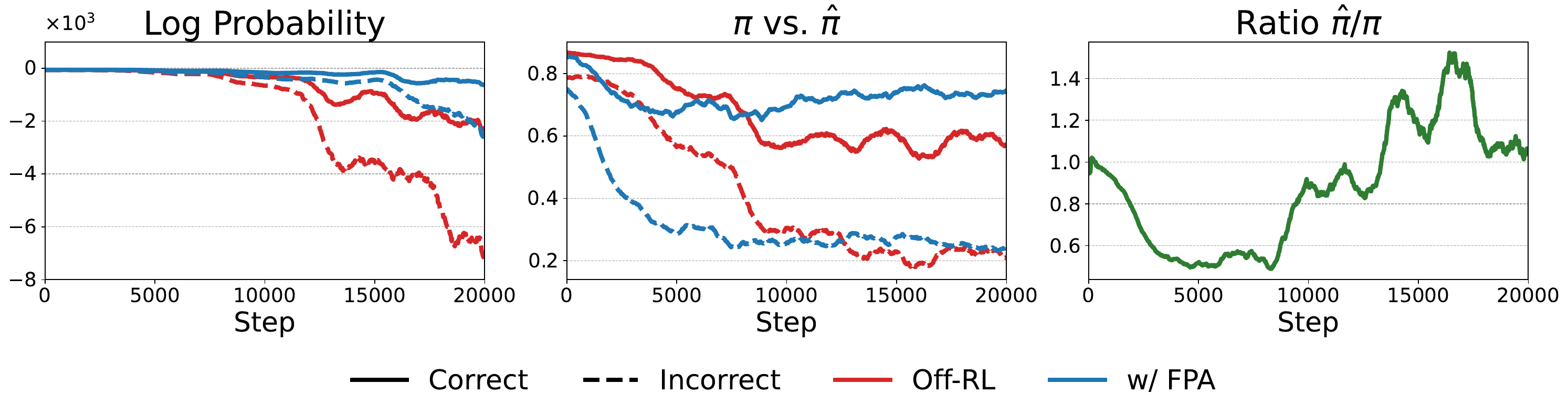}
    \caption{\textbf{Training Dynamics.} (\textbf{Left}) Log-probability difference $\log \pi_\theta (y\mid x) - \log\pi_\text{ref}(y\mid x)$ during training for correct and incorrect trajectories for Off-RL and \method{}. (\textbf{Center}) The policy probability used for gradients during training, $\pi$ for offline RL and $\hat{\pi}$ for \method{}. (\textbf{Right}) The ratio $\hat{\pi}$ from \method{} and $\pi$ from offline RL for incorrect trajectories during training.}
    \label{fig:dynamics}
    \vspace{-10pt}
\end{figure}

\paragraph{Training Dynamics of \method{}.}
To understand the empirical effects of \method{}, we analyze its training dynamics in detail. As shown in Figure~\ref{fig:dynamics} (left), under Off-RL, the log-probabilities of \emph{both} correct and incorrect trajectories decrease during training due to gradient entanglement: excessive penalization of incorrect trajectories also suppresses correct ones. Notably, similar behavior has been observed when applying the widely used offline RL algorithm DPO to mathematical reasoning~\citep{pal2024smaug}. In contrast, \method{} largely preserves the log-probabilities of correct trajectories throughout training, indicating that shared useful tokens are not excessively penalized.

Figure~\ref{fig:dynamics} (center) illustrates how this behavior arises from the proactive nature of \method{}. We compare the probability weights used for gradient updates---$\pi$ under Off-RL and $\hat{\pi}$ under \method{}---for correct and incorrect trajectories. For incorrect trajectories, $\hat{\pi}$ decreases early, acting as a proactive regularizer that dampens incorrect gradient updates before they over-penalize shared useful tokens. By contrast, $\pi$ under Off-RL decreases only after the trajectory log-probabilities have already fallen, causing the regularization to arrive too late to prevent the suppression of correct trajectories. Figure~\ref{fig:dynamics} (right) further illustrates this effect through the ratio $\hat{\pi}/\pi$ for incorrect trajectories. Although the two weights are initially similar, $\hat{\pi}$ becomes substantially smaller than $\pi$ early in training, confirming its role as an \emph{early brake} on incorrect gradients. As training progresses, the current-policy weight $\pi$ eventually decreases, and the ratio $\hat{\pi}/\pi$ moves back toward 1, but only late in training, when the benefit of proactive regularization is diminished.

Figure~\ref{fig:dynamics} (center) also reveals a contrasting pattern for correct trajectories. Under Off-RL, as $\pi$ decreases during training, the gradient weights assigned to correct samples shrink, limiting the exploitation of high-probability correct trajectories that Off-RL is intended to reinforce (see \S~\ref{sec:prelim_Off-RL}). Under \method{}, $\hat{\pi}$ for correct trajectories shows a mild initial decrease, likely reflecting early gradient entanglement, but subsequently recovers and increases. This recovery, which is absent under Off-RL, suggests that \method{} not only prevents excessive suppression of correct trajectories but also restores their reinforcing signal as training proceeds. Consequently, \method{} maintains larger gradient weights for correct trajectories and preserves their learning signal throughout training.

\begin{figure}[!ht]
\centering
    \includegraphics[width=\textwidth]{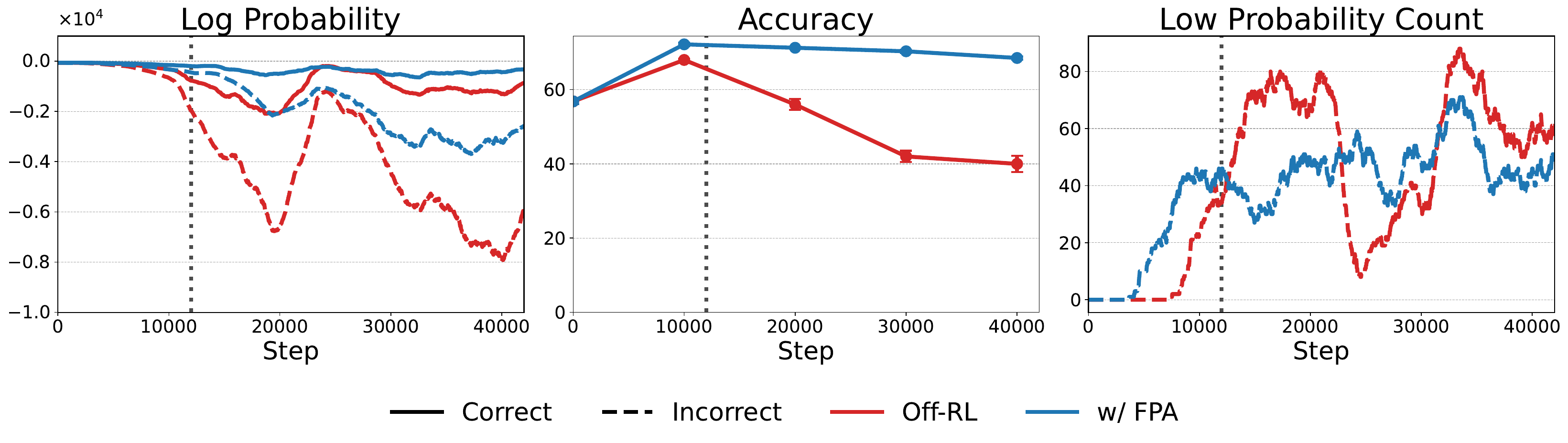}
    \caption{\textbf{Training dynamics under prolonged training.} (\textbf{Left}) Log-probabilities $\log\pi_\theta(y\mid x) - \log\pi_\text{ref}(y\mid x)$ for correct and incorrect trajectories under Off-RL and \method{}. (\textbf{Center}) MATH500 accuracy throughout training. (\textbf{Right}) The number of times $\pi$ and $\hat{\pi}$ fall below 0.2 for incorrect trajectories, indicating strong regularization of incorrect gradients.}
    \label{fig:prolonged}
\end{figure}

\subsection{Further Analysis}
\label{sec:exp_further}

\paragraph{Model Degradation.}
Over-penalization of shared tokens not only limits performance gains but can also lead to model degradation. As shown in Figure~\ref{fig:prolonged} (left), prolonged training under Off-RL causes the log-probabilities of both correct and incorrect trajectories to continuously decrease. Notably, the log-probabilities of correct trajectories under Off-RL eventually fall below those of incorrect trajectories under \method{}, highlighting the severity of over-penalization. In contrast, \method{} keeps the log-probabilities of correct trajectories substantially more stable throughout training. As shown in Figure~\ref{fig:prolonged} (center), this over-penalization is accompanied by a clear performance drop at around 14K steps, whereas \method{} remains relatively stable.

To further analyze this effect, Figure~\ref{fig:prolonged} (right) shows how frequently $\pi$ and $\hat{\pi}$ fall below a threshold of 0.2, capturing how often gradients from incorrect trajectories are strongly regularized. Under \method{}, this frequency rises before visible performance degradation, reflecting its proactive nature. Under Off-RL, it rises later and more abruptly---only after performance has already begun to collapse---making the regularization too late to prevent degradation. This suggests that $\hat{\pi}$ does not merely scale $\pi$ uniformly but instead adaptively suppresses potentially destabilizing updates before they become harmful. Similar performance collapse under extended training has been reported for DPO by \citet{pal2024smaug}.

\begin{wrapfigure}{r}{0.48\columnwidth}
    \vspace{-1.0em}
    \centering
    \includegraphics[width=\linewidth]{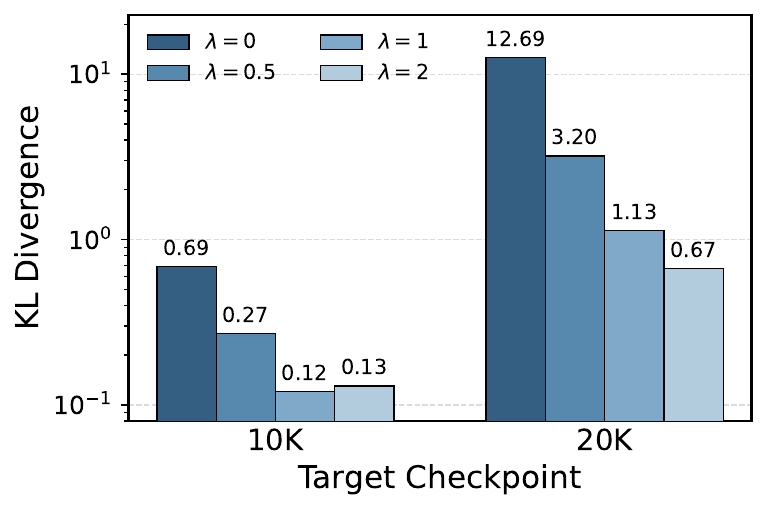}
    \caption{\textbf{Validating the future-policy approximation.}
    KL divergence between policies extrapolated from the 5K checkpoint and the actual policies at 10K and 20K steps. $\lambda=0$ corresponds to the unmodified 5K policy. Lower is better.}
    \label{fig:future_policy_kl}
    \vspace{-1.0em}
\end{wrapfigure}

\paragraph{Validating the Future Policy.}
We directly validate whether \method{} provides a meaningful approximation of the future policy. We anchor Qwen3-4B-Base at the 5K-step checkpoint, extrapolate the policy relative to the reference policy (0K), and compare the resulting estimates with the actual policies obtained at 10K and 20K steps. We measure KL divergence using MATH500 prompts with 8 samples per prompt, following the same sampling protocol as our evaluation. As shown in Figure~\ref{fig:future_policy_kl}, the extrapolated policies are consistently closer to the actual future checkpoints than the unmodified 5K policy. At 20K steps, for example, the KL divergence decreases from 12.69 for the 5K anchor to 0.67 with $\lambda=2$. Moreover, the optimal extrapolation strength increases with the prediction horizon: $\lambda=1$ is closest to the 10K checkpoint, whereas $\lambda=2$ is closest to the 20K checkpoint. These results provide direct evidence that the extrapolation captures subsequent policy movement during Off-RL training.

\paragraph{Algorithm Ablation.} 
We conduct a series of ablations to examine the design choices behind \method{}. As shown in Table~\ref{tab:ablation_results}, full \method{} yields the best overall performance.

\begin{table*}[!tb]
\centering
\caption{\textbf{Algorithm ablation results} on Qwen3-4B-Base. All results are reported as Pass@1 accuracy. The best performance is \textbf{bolded}, and the second best is \underline{underlined}.}
\small
\setlength{\tabcolsep}{3pt}
\renewcommand{\arraystretch}{1.15}
\begin{tabularx}{\textwidth}{l *{8}{C}}
\toprule
\textbf{Method}
& \textbf{GSM8K}
& \textbf{MATH500}
& \textbf{\mbox{MATH-P}}
& \textbf{GaoKao}
& \textbf{Olym.}
& \textbf{AMC}
& \textbf{AIME}
& \textbf{Avg.} \\
\midrule
Base   & 75.52 & 56.83 & 33.18 & 43.96 & 25.40 & 29.78 & 6.60 & 38.75 \\
Off-RL & 86.80 & 66.70 & 41.42 & 52.76 & 30.72 & 36.91 & 8.50 & 46.26 \\
\hspace{0.5em}w/ Low LR & 80.85 & 60.40 & 38.04 & 48.96 & 29.13 & 34.22 & 7.50 & 42.73 \\
\hspace{0.5em}w/ RS & 86.90 & 65.90 & 39.90 & 53.80 & 30.30 & 30.60 & 8.20 & 45.09 \\
\hspace{0.5em}w/ Cor. Only & 40.11 & 31.50 & 14.34 & 22.40 & 9.17 & 11.30 & 1.11 & 18.56 \\
\hspace{0.5em}w/ Incor. Only & \underline{88.51} & 68.88 & 40.37 & 56.04 & 32.96 & 39.26 & 9.24 & 47.89 \\
\hspace{0.5em}w/ KL & 76.61 & \underline{71.73} & \underline{48.19} & \underline{56.79} & \textbf{35.78} & \underline{39.51} & \textbf{10.28} & \underline{48.41} \\
\rowcolor{rowhl}
\hspace{0.5em}\textbf{w/ \method{}} & \textbf{91.60} & \textbf{74.50} & \textbf{48.40} & \textbf{59.60} & \underline{35.40} & \textbf{42.80} & \underline{10.10} & \textbf{51.77} \\
\bottomrule
\end{tabularx}
\label{tab:ablation_results}
\end{table*}

\begin{figure}[!t]
\centering
    \includegraphics[width=\textwidth]{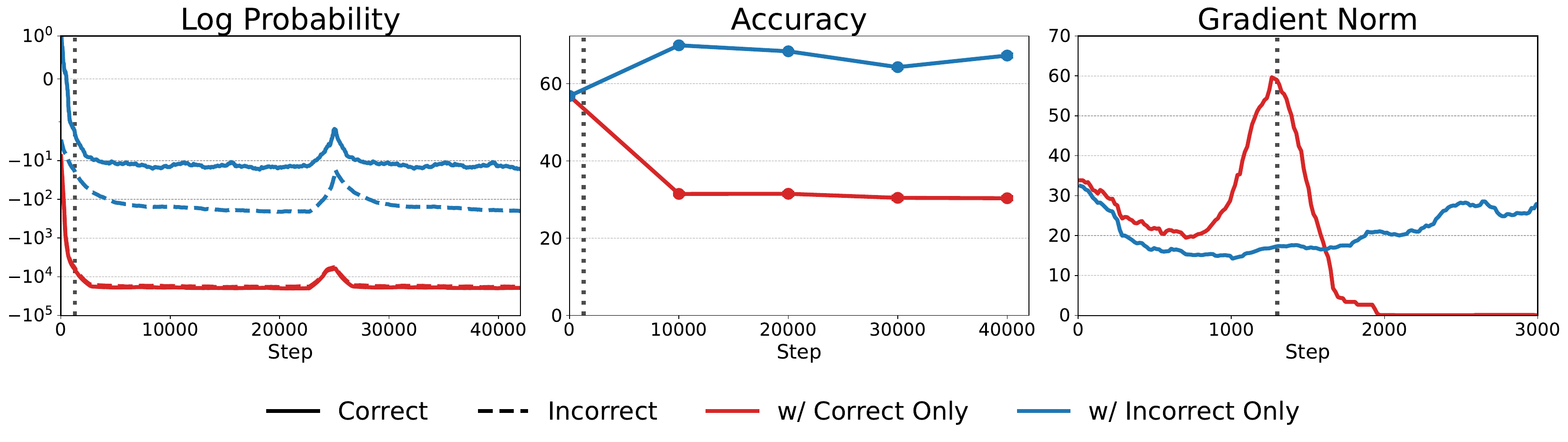}
    \caption{\textbf{Targeted \method{} ablations.} (\textbf{Left}) Log-probabilities $\log\pi_\theta(y \mid x) - \log\pi_\text{ref}(y\mid x)$ for \textit{w/ Cor. Only} and \textit{w/ Incor. Only}. (\textbf{Center}) MATH500 accuracy throughout training. (\textbf{Right}) Gradient norm throughout training.}
    \label{fig:pos}
    \vspace{-18pt}
\end{figure}

\begin{itemize}[leftmargin=*]
    \item \textbf{Learning rate decay.}
    To test whether \method{}'s gains simply arise from smaller gradient steps, we make the cosine learning-rate schedule decay more aggressively, reducing the learning rate to 80\% of the original Off-RL schedule (\textit{w/ Low LR}). As shown in Table~\ref{tab:ablation_results}, this stronger decay only limits learning.

    \item \textbf{Reward Shaping.}
    We also test a simpler way to reduce the influence of incorrect trajectories by statically weakening their negative rewards. Specifically, we keep $R=+1$ for correct trajectories and set $R=-0.5$ for incorrect trajectories (\textit{w/ RS}). As shown in Table~\ref{tab:ablation_results}, this also does not improve upon Off-RL. Together, these results suggest that \method{}'s gains do not come from simply reducing the update magnitude, but from its proactive, trajectory-dependent reweighting. We report the full reward-scale sweep in \S~\ref{app:baseline_tuning}.

    \item \textbf{Targeted \method{}.}
    To pinpoint where \method{}'s gains come from, we apply \method{} separately to correct and incorrect trajectories. As shown in Table~\ref{tab:ablation_results}, applying \method{} only to incorrect trajectories (\textit{w/ Incor. Only}) substantially improves over Off-RL but remains below the full \method{}, indicating that regularizing gradients from incorrect trajectories accounts for a major part of the gain, while reweighting correct trajectories also contributes. In contrast, applying \method{} only to correct trajectories (\textit{w/ Cor. Only}) leads to model collapse. Without regularization of gradients from incorrect trajectories, the log-probabilities of correct trajectories decrease rather than increase during training (see \S~\ref{sec:exp_math}); consequently, $\hat{\pi}$, which approximates the future policy, predicts even lower probabilities and further suppresses gradients from correct trajectories. Figure~\ref{fig:pos} further supports this interpretation. In the left and center panels, \textit{w/ Cor. Only} shows a steep decrease in the log-probabilities of both correct and incorrect trajectories and a corresponding performance collapse, whereas \textit{w/ Incor. Only} remains stable. The right panel shows a sharp gradient-norm spike at the point of collapse, consistent with recent observations that model collapse is accompanied by gradient-norm spikes~\citep{deng2025grpo}.

    \item \textbf{KL regularization.}
    A common strategy for stabilizing RL is to regularize the policy toward a reference model using a KL-divergence penalty~\citep{schulman2015trust}. We therefore augment Off-RL with a KL penalty. As shown in Table~\ref{tab:ablation_results} (\textit{w/ KL}), this improves over vanilla Off-RL but remains substantially below \method{}. This suggests that KL regularization alone does not provide the same benefit as \method{}'s proactive reweighting. Further details can be found in \S~\ref{app:kl_ablation} and we additionally report the full sweep over KL regularization strengths in \S~\ref{app:baseline_tuning}.
\end{itemize}

\paragraph{Step-Level Supervision.}
\begin{table*}[!t]
\centering
\caption{\textbf{Results on Step-DPO preference data.} Results are obtained on Math-Step-DPO-10K~\citep{lai2024step} with Qwen3-4B-Base. All results are Pass@1 accuracy. The best performance is \textbf{bolded}, and the second-best performance is \underline{underlined}.}
\small
\setlength{\tabcolsep}{3pt}
\renewcommand{\arraystretch}{1.15}
\begin{tabularx}{\textwidth}{l *{8}{C}}
\toprule
\textbf{Method} & \textbf{GSM8K} & \textbf{MATH500} & \textbf{MATH-P} & \textbf{GaoKao} & \textbf{Olym.} & \textbf{AMC} & \textbf{AIME} & \textbf{Avg.} \\
\midrule
Base    & 75.52 & 56.83 & 33.18 & 43.96 & 25.40 & 29.78 & 6.60 & 38.75 \\
DPO     & 77.94 & 64.85 & \underline{41.42} & 49.94 & 30.17 & \textbf{37.42} & 7.78 & 44.22 \\
Off-RL  & \underline{85.13} & \underline{65.28} & 39.16 & \underline{52.14} & \underline{30.57} & 33.53 & \underline{8.08} & \underline{44.84} \\
\rowcolor{rowhl}
\hspace{0.5em}\textbf{w/ \method{}} & \textbf{88.13} & \textbf{68.80} & \textbf{42.61} & \textbf{55.29} & \textbf{32.35} & \underline{33.84} & \textbf{9.86} & \textbf{47.27} \\
\bottomrule
\end{tabularx}
\label{tab:stepdpo}
\vspace{-8pt}
\end{table*}

Another approach to mitigating gradient entanglement in mathematical reasoning is to use step-level supervision, which localizes feedback to individual reasoning steps rather than entire trajectories. We train on Math-Step-DPO-10K~\citep{lai2024step} using Qwen3-4B-Base. As shown in Table~\ref{tab:stepdpo}, \method{} achieves the best average accuracy, outperforming DPO and Off-RL by 3.05\% and 2.43\%, respectively. This suggests that step-level supervision does not fully eliminate the need for proactive gradient reweighting, as correct and incorrect steps may still share substantial token overlap. Thus, \method{} is complementary to step-level supervision. Notably, the lower absolute performance compared with Table~\ref{tab:main_results} may partly reflect the fact that Math-Step-DPO-10K is not self-generated by the model, consistent with prior findings on the importance of self-generated reasoning data~\citep{setlur2024rl}.

\begin{wrapfigure}{r}{0.4\columnwidth}
  \vspace{-2.5em}
  \centering
  \begin{minipage}[t]{0.4\columnwidth}
    \centering
    \includegraphics[width=\linewidth]{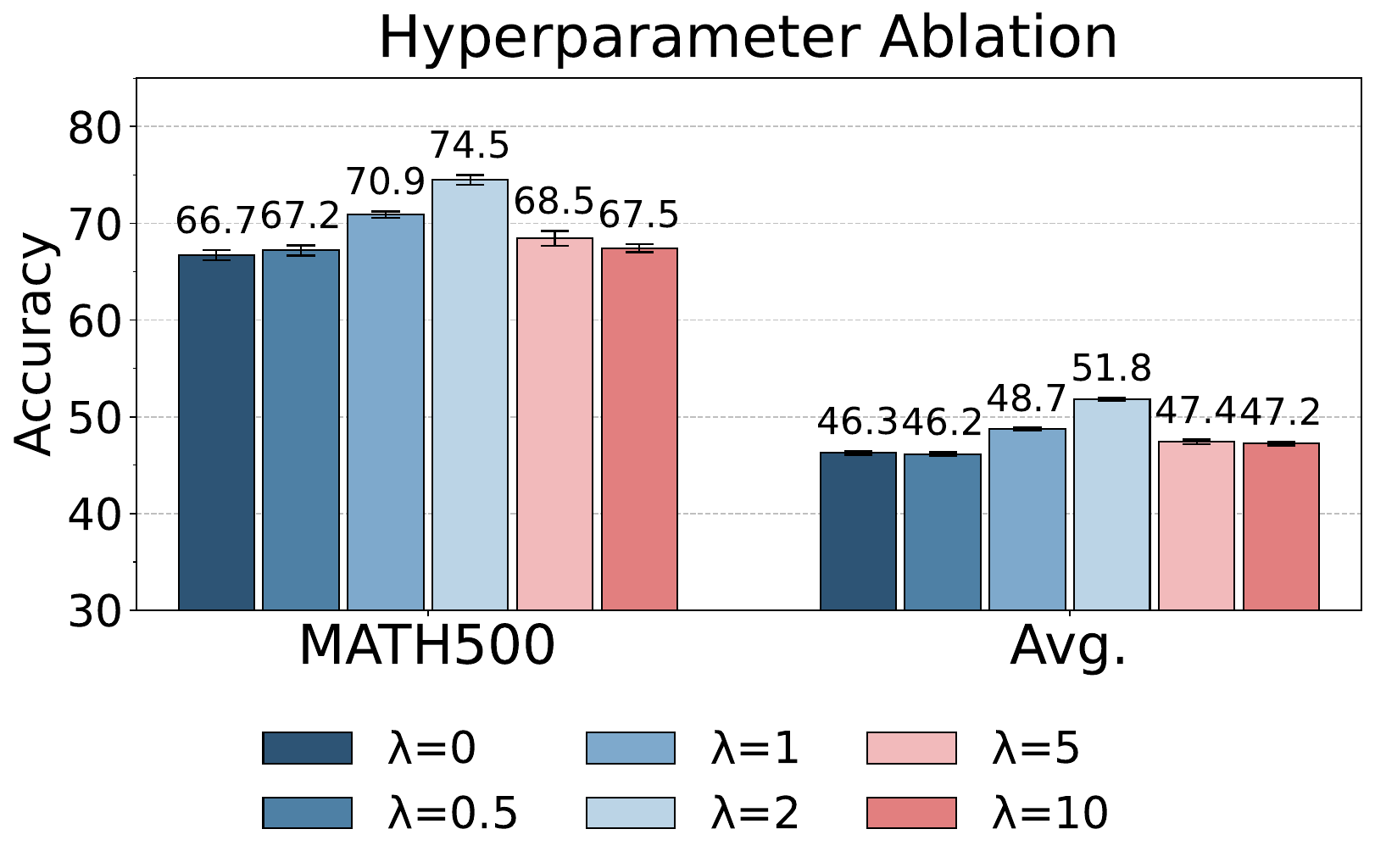}
    \captionof{figure}{\textbf{$\lambda$ Sensitivity} on MATH500 and
    average accuracy of \method{} across
    $\lambda$.}
    \label{fig:lambda}
  \end{minipage}
  \\[0.5em]
  \begin{minipage}[t]{0.4\columnwidth}
    \centering
    \includegraphics[width=\linewidth]{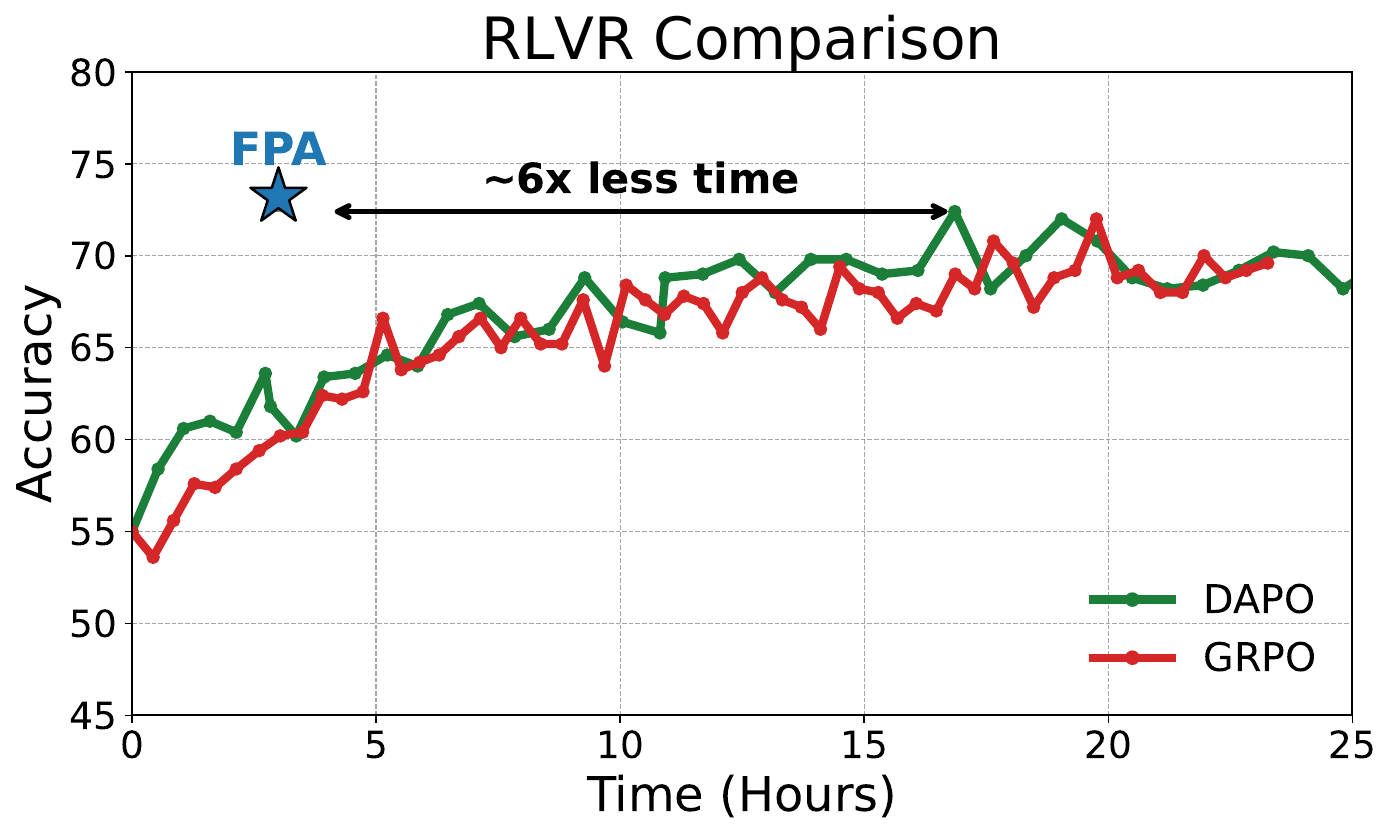}
    \captionof{figure}{\textbf{Comparison to RLVR} with MATH500 accuracy
    over wall-clock training time for \method{}, DAPO and GRPO.}
    \label{fig:grpo}
  \end{minipage}
  \vspace{-30pt}
\end{wrapfigure}

\paragraph{Hyperparameter Ablation.}
\label{sec:exp_hyperparameter}
We analyze the sensitivity of \method{} to the extrapolation hyperparameter $\lambda$ by ablating over $\lambda \in \{0.5, 1, 2, 5, 10\}$. Figure~\ref{fig:lambda} reports MATH500 and average accuracy. \method{} consistently outperforms Off-RL ($\lambda=0$) across a wide range of values, with $\lambda \in \{1, 2\}$ performing best. Notably, even an excessively large value of $\lambda=10$ remains above Off-RL, although the gains diminish as overly distant extrapolation can over-regularize useful updates. This broad range of effective values suggests that \method{} is not highly sensitive to the precise choice of $\lambda$. Validation-set results for $\lambda$ selection and further ablations are provided in \S~\ref{app:hyperparameter}.

\paragraph{Comparison with RLVR.}
\label{sec:exp_grpo}
We compare \method{} against the widely used RLVR methods GRPO~\citep{shao2024deepseekmath} and DAPO~\citep{yu2025dapo}, a state-of-the-art variant, in terms of both accuracy and wall-clock efficiency. As shown in Figure~\ref{fig:grpo}, \method{} reaches $73\%$ accuracy in approximately 3 hours, including 2 hours of training and 1 hour of data generation on 2 H200 GPUs, whereas DAPO requires around 17 hours to reach comparable performance, corresponding to a roughly $6\times$ reduction in wall-clock time (see \S~\ref{app:dapo}). Thus, under this hardware-matched setting, \method{} provides a substantially more efficient alternative to online RLVR. Moreover, offline data generation is a one-time cost: the same data can be reused across hyperparameter sweeps and algorithm comparisons without additional generation, further reducing the cost of iterative experimentation.

\subsection{Coding Experiment Results}
\label{sec:exp_coding}

\begin{wraptable}{r}{0.5\textwidth}
\vspace{-1.25em}
\centering
\caption{\textbf{Code-generation results} across two model families. All results are Pass@1 accuracy. The best performance is \textbf{bolded}, and the second best is \underline{underlined}.}
\label{tab:coding_main}
\small
\setlength{\tabcolsep}{4pt}
\begin{tabular}{l|cccc}
\toprule
\textbf{Method} & \textbf{MBPP+} & \textbf{BCB} & \textbf{LCB} & \textbf{Avg.} \\
\midrule
\multicolumn{5}{l}{\textit{Qwen3-4B-Base}} \\
Base
& 50.08 & 34.26 & 11.53 & 31.96 \\
DPO
& \underline{60.15} & 38.62 & 22.20 & 40.32 \\
Off-RL
& 60.05 & \underline{38.90} & \textbf{23.41} & \underline{40.79} \\
\rowcolor{gray!15}
\textbf{w/ \method{}}
& \textbf{61.74} & \textbf{39.95} & \underline{23.19} & \textbf{41.63} \\
\midrule
\multicolumn{5}{l}{\textit{Llama-3.2-3B-Instruct}} \\
Base
& 47.09 & 13.30 & 9.43 & 23.27 \\
DPO
& \textbf{48.94} & 19.09 & 10.29 & \underline{26.11} \\
Off-RL
& 46.30 & \underline{20.91} & \underline{10.86} & 26.02 \\
\rowcolor{gray!15}
\textbf{w/ \method{}}
& \underline{48.41} & \textbf{21.82} & \textbf{12.00} & \textbf{27.41} \\
\bottomrule
\end{tabular}
\vspace{-0.5em}
\end{wraptable}

We further evaluate \method{} on code generation to examine whether its benefits extend beyond mathematical reasoning. Similar to mathematical reasoning, correct and incorrect programs often share substantial portions of their token sequences, including function signatures, control flow, and partially correct implementations. As shown in Table~\ref{tab:coding_main}, \method{} achieves the highest average performance on both models. On Qwen3-4B-Base, \method{} reaches an average Pass@1 of 41.63, compared with 40.79 for Off-RL and 40.32 for DPO. On Llama-3.2-3B-Instruct, \method{} similarly achieves the best average performance of 27.41, with the strongest results on BCB and LCB. In both cases, all training methods improve average performance over the corresponding base model, while \method{} provides the strongest overall performance. We further verify across three random seeds in \S~\ref{app:multiseed}.

\section{Related Work}
\subsection{LLM Reasoning}
Recent advances in LLM reasoning have increasingly relied on allocating more computation at inference time. Chain-of-Thought prompting first showed that generating intermediate reasoning steps can improve performance on complex tasks~\citep{cot2022}. Follow-up work extended this through diverse sampling and aggregation~\citep{wang2023selfconsistency}, iterative refinement~\citep{madaan2023self}, and explicit search over reasoning paths~\citep{yao2023tree}. Verifier-based methods further use outcome or process supervision to select better solutions or intermediate steps~\citep{lightman2024let}. More recently, these ideas have developed into \emph{test-time scaling}, where additional inference compute is allocated through longer reasoning, repeated sampling, search, or verification~\citep{snell2025scaling}. Subsequent methods directly control reasoning length through budget forcing~\citep{muennighoff2025s1} or adaptively allocate inference-time computation~\citep{song2026thinkbrake, yang2026dynamic}. Related work also explores confidence-aware compute allocation and latent-space reasoning as alternatives to simply generating longer chains~\citep{geiping2026scaling}. Together, these works establish inference-time computation through Chain-of-Thought reasoning as a central component of modern LLM reasoning.

\subsection{Learning to Reason}
\paragraph{Supervised and Preference Learning.}
Early approaches to learning reasoning capabilities relied primarily on supervised fine-tuning over successful reasoning trajectories, using iterative generation, filtering, and fine-tuning on successful model-generated solutions. These approaches demonstrated that models can improve by learning from their own reasoning traces~\citep{zelikman2022star,yuan2023rft,gulcehre2023rest}. However, training only on successful trajectories fails to leverage information from unsuccessful reasoning attempts. Preference optimization provides a natural offline alternative that learns from both preferred and dispreferred responses. DPO~\citep{rafailov2024dpo} directly optimizes preference pairs without an explicit reward model, followed by variants such as KTO~\citep{Ethayarajh2024KTO} and SimPER~\citep{xiao2025simper}, which simplify or modify the preference-learning objective. However, applying preference optimization to long-horizon reasoning remains challenging, particularly because of \textit{gradient entanglement}, where correct and incorrect trajectories share substantial portions of their reasoning~\citep{lai2024step,yuan2025common}. This has motivated reasoning-specific approaches such as step-level preference optimization~\citep{lai2024step} and objectives that modify the treatment of preferred and dispreferred gradients~\citep{pal2024smaug, pang2024iterativerpo}.

\paragraph{Online Reinforcement Learning.}
Reinforcement Learning from Verifiable Rewards (RLVR) has become one of the most successful approaches for eliciting reasoning, typically assigning trajectory-level rewards based on answer correctness. GRPO~\citep{shao2024deepseekmath} lies at the core of this paradigm, and subsequent methods have improved its efficiency and stability through techniques such as dynamic sampling, modified clipping, and importance-sampling schemes~\citep{yu2025dapo,chen2025minimax,wang2025aspo,zheng2025gspo}. More recent work further reduces group-level sampling overhead~\citep{xu2026single,choi2026your}. Despite their strong performance, these methods repeatedly generate fresh trajectories from the evolving policy, making training computationally expensive and potentially unstable~\citep{liu2025rlcollapse,zhang2026mismatch}. A related direction that has recently gained popularity is On-Policy Distillation (OPD)~\citep{lu2025onpolicydistillation,gu2024minillm,agarwal2024onpolicy}, where the student generates its own trajectories and learns from dense token-level supervision provided by a stronger teacher, often through an RL-style objective~\citep{ko2026scaling, oh2026kl}. Despite its strong performance, OPD still requires costly online generation and access to a stronger teacher, which may not always be available.

\paragraph{Offline Reinforcement Learning.} 
An orthogonal line of work avoids repeated online generation by learning from fixed or previously generated reasoning trajectories, thereby reducing rollout costs and improving data efficiency. Recent work has begun to demonstrate the viability of simple policy-gradient and soft Bellman objective-based methods for mathematical reasoning~\citep{lu2026pcl, wang2025offline}. Related work, such as $A^\star$-PO~\citep{brantley2025accelerating}, uses offline samples to estimate an optimal value function before performing lightweight on-policy optimization. More broadly, recent off-policy methods reuse lagged, external, or replayed trajectories to reduce online sampling costs~\citep{yan2026luffy,liang2026squeeze}. Nevertheless, direct offline policy-gradient RL on a fixed reasoning dataset remains comparatively underexplored, particularly in gradient-entangled settings. \method{} addresses this setting through proactive future-policy reweighting.

\section{Conclusion}
We introduced \textbf{Future Policy Approximation (\method{})}, a simple and computationally lightweight approach for stabilizing offline policy-gradient training for LLM reasoning. \method{} uses a logit-space-extrapolated future policy to proactively reweight trajectory gradients, mitigating over-penalization caused by gradient entanglement between correct and incorrect reasoning trajectories. Across three models, seven mathematical reasoning benchmarks, and three code-generation benchmarks, \method{} improves over strong offline baselines and remains stable under prolonged training. We further validate its effectiveness under step-level supervision, across multiple random seeds, and through direct analysis of the future-policy approximation. Under our compute-matched setting, \method{} also achieves performance comparable to online RLVR at substantially lower training cost, making it a practical approach for efficient offline reasoning optimization.


\bibliography{main}
\bibliographystyle{tmlr}

\appendix
\clearpage
\section{Extended Theory}
\subsection{Decomposing SimPER to RL}
\label{app:simper_Off-RL}
In this section, we show that SimPER \citep{xiao2025simper} can be understood as a special case of the Off-RL objective (Eq.~\ref{eq:RL}). SimPER is a preference learning method based on inverse-perplexity that does not require a reference model or additional hyperparameters:
\begin{align}
    \mathcal{L}_\text{SimPER}(\theta)
    = - \mathbb{E}_{(x, y_w, y_l) \sim \mathcal{D}}
    \bigg[ \exp\!\Big(\tfrac{1}{|y_w|}\log \pi_\theta(y_w \mid x)\Big)
          - \exp\!\Big(\tfrac{1}{|y_l|}\log \pi_\theta(y_l \mid x)\Big) \bigg]
\end{align}

\paragraph{Notation.}
Let $\mathcal{X}$ be the input space and $\mathcal{Y}^\ast$ the set of all finite token sequences.
For $y\in\mathcal{Y}^\ast$, let $|y|$ denote its length.
We denote the language model policy by $\pi_\theta$, with parameters $\theta\in\Theta$.
Given an input $x\in\mathcal{X}$, a response $y\in\mathcal{Y}^\ast$ is generated from the policy, written as
\[
y \sim \pi_\theta(\cdot \mid x).
\]
The policy factorizes as
\[
\pi_\theta(y\mid x)=\prod_{t=1}^{|y|}\pi_\theta\!\big(y_t \mid x, y_{<t}\big).
\]

For each input $x$, we use a fixed reference policy $\pi_{\mathrm{ref}}$ to generate a pairwise preference set
\[
S_x=\{y_w(x),\, y_l(x)\}\subset\mathcal{Y}^\ast,
\]
where $y_w$ denotes the correct response and $y_l$ the incorrect response. We denote the training dataset by
\[
\mathcal{D}=\{(x, S_x)\}_{i=1}^N .
\]
Throughout, we assume $\pi_\theta(y\mid x)>0$ for all $y\in S_x$ so that $\log \pi_\theta(y\mid x)$ is well-defined.

To map this to the reinforcement learning formulation, we define a masked reward $R(x,y)$ such that $R(x,y_w) = +1$ for correct responses and $R(x,y_l) = -1$ for incorrect responses. Based on our notation, we can now rewrite the Off-RL loss as:
\begin{equation}\label{eq:app:rl}
J_{\mathrm{Off-RL}}(\theta)
:= \mathbb{E}_{(x,S_x)\sim \mathcal D}\!\left[\sum_{y\in S_x} \pi_\theta(y\mid x)^{\frac{1}{|y|}}\,R(x,y)\right].
\end{equation}

We can now state the following proposition showing that SimPER implicitly optimizes the Off-RL objective.

\begin{proposition}[SimPER aligns with Off-RL]
The gradient of SimPER's objective equals the gradient of Off-RL as follows:
\[
\nabla_\theta J_{\mathrm{SimPER}}(\theta) = \nabla_\theta J_{\mathrm{Off-RL}}(\theta).
\]
\end{proposition}

\begin{proof}
We start from SimPER's objective function:
\begin{align}
J_{\mathrm{SimPER}}(\theta) &= -\mathcal{L}_{\mathrm{SimPER}}(\theta) \nonumber\\
&= \mathbb{E}_{(x,S_x)\sim \mathcal D}\!\left[
\exp\!\Big(\tfrac{1}{|y_w|}\log\pi_\theta(y_w\mid x)\Big)
-\exp\!\Big(\tfrac{1}{|y_l|}\log\pi_\theta(y_l\mid x)\Big)
\right].
\end{align}
Let
\[
f_\theta(y\mid x)
:= \exp\!\Big(\tfrac{1}{|y|}\log\pi_\theta(y\mid x)\Big).
\]
Using this notation, SimPER's objective can be written as:
\[
J_{\mathrm{SimPER}}(\theta) = \mathbb{E}_{(x,S_x)\sim \mathcal D}\!\left[
f_\theta(y_w\mid x) - f_\theta(y_l\mid x)\right].
\]
Taking the derivative with respect to $\theta$, we obtain: 
\begin{align}\label{eq:simper-obj-grad}
\nabla_\theta J_{\mathrm{SimPER}}(\theta)
&= \mathbb{E}_{(x,S_x)\sim \mathcal D}\!\left[
\nabla_\theta f_\theta(y_w\mid x) - \nabla_\theta f_\theta(y_l\mid x)\right] \nonumber\\
&= \mathbb{E}_{(x,S_x)\sim \mathcal D}\!\Big[
\pi_\theta(y_w\mid x)^{\frac{1}{|y_w|}}\,|y_w|^{-1}\,\nabla_\theta \log\pi_\theta(y_w\mid x) \nonumber\\
&\qquad\qquad\qquad\quad -\pi_\theta(y_l\mid x)^{\frac{1}{|y_l|}}\,|y_l|^{-1}\,\nabla_\theta \log\pi_\theta(y_l\mid x)\Big].
\end{align}
Since $R(x,y_w) = +1$ and $R(x,y_l) = -1$, we can rewrite Eq.~\ref{eq:simper-obj-grad} as:
\begin{equation}\label{eq:simper-grad}
\nabla_\theta J_{\mathrm{SimPER}}(\theta) = \mathbb{E}_{(x,S_x)\sim \mathcal D}\!\left[
\sum_{y\in S_x}\pi_\theta(y\mid x)^{\frac{1}{|y|}}\,|y|^{-1}\,
\nabla_\theta \log\pi_\theta(y\mid x)\,R(x,y)\right].
\end{equation}

On the other hand, recall the Off-RL objective Eq.~\ref{eq:app:rl}:
\begin{equation}\label{eq:ln-objective}
J_{\mathrm{Off-RL}}(\theta) = \mathbb{E}_{(x,S_x)\sim \mathcal D}\!\left[\sum_{y\in S_x} \pi_\theta(y\mid x)^{\frac{1}{|y|}}\,R(x,y)\right].
\end{equation}
Differentiating both sides of Eq.~\ref{eq:ln-objective} with respect to $\theta$ yields:
\begin{align}
\nabla_\theta J_{\mathrm{Off-RL}}(\theta) 
&= \mathbb{E}_{(x,S_x)\sim \mathcal D}\!\left[\sum_{y\in S_x} \nabla_\theta\left(\pi_\theta(y\mid x)^{\frac{1}{|y|}}\right)\,R(x,y)\right] \nonumber\\
&= \mathbb{E}_{(x,S_x)\sim \mathcal D}\!\left[\sum_{y\in S_x} \pi_\theta(y\mid x)^{\frac{1}{|y|}}\,|y|^{-1}\,\nabla_\theta \log\pi_\theta(y\mid x)\,R(x,y)\right]. \label{eq:ln-grad}
\end{align}
Since the right-hand sides of Eq.~\ref{eq:simper-grad} and Eq.~\ref{eq:ln-grad} are identical, the following equality holds:
\begin{equation}
\nabla_\theta J_{\mathrm{SimPER}}(\theta) = \nabla_\theta J_{\mathrm{Off-RL}}(\theta),
\end{equation}
which completes the proof.
\end{proof}

\subsection{Limitations of DPO with Deterministic Preferences}
\label{app:dpo_limit}
Direct Preference Optimization (DPO) may not be robust in mathematical reasoning tasks due to its reliance on the Bradley-Terry model \citep{Bradley1952}. Under this model, DPO assumes that preference pairs are selected based on an unknown reward function $r^*(x,y)$, where the probability that response $y_w$ is preferred over $y_l$ follows:
\begin{align}
&p^*(y_w \succ y_l \mid x) \nonumber \\
&\quad = \frac{\exp(r^*(x, y_w))}{\exp(r^*(x, y_w)) + \exp(r^*(x, y_l))} \nonumber \\
&\quad = \sigma\!\big(r^*(x, y_w) - r^*(x, y_l)\big).
\end{align}

This assumption becomes problematic in mathematical reasoning tasks, where preference pairs consist of clearly correct and incorrect answers, yielding deterministic preferences: $p^*(y_w \succ y_l \mid x) = 1$. As \citet{azar2024general} and \citet{fisch2024robust} show, deterministic preferences require $r^*(y_w) - r^*(y_l) \to \infty$ in the Bradley-Terry model, forcing $\pi_{\theta^*}(y_l \mid x) = 0$ regardless of the KL regularization strength $\beta$. Since mathematical reasoning trajectories often share a large number of common tokens between correct and incorrect sequences, this over-penalization of the incorrect sequences can lead to a degradation of the model's overall performance.

\clearpage
\section{Experiment Details}
\label{app:exp_details}

\subsection{Pilot Experiment Details}
\label{app:pilot_details}
\begin{figure}[!ht]
\centering
    \includegraphics[width=0.6\textwidth]{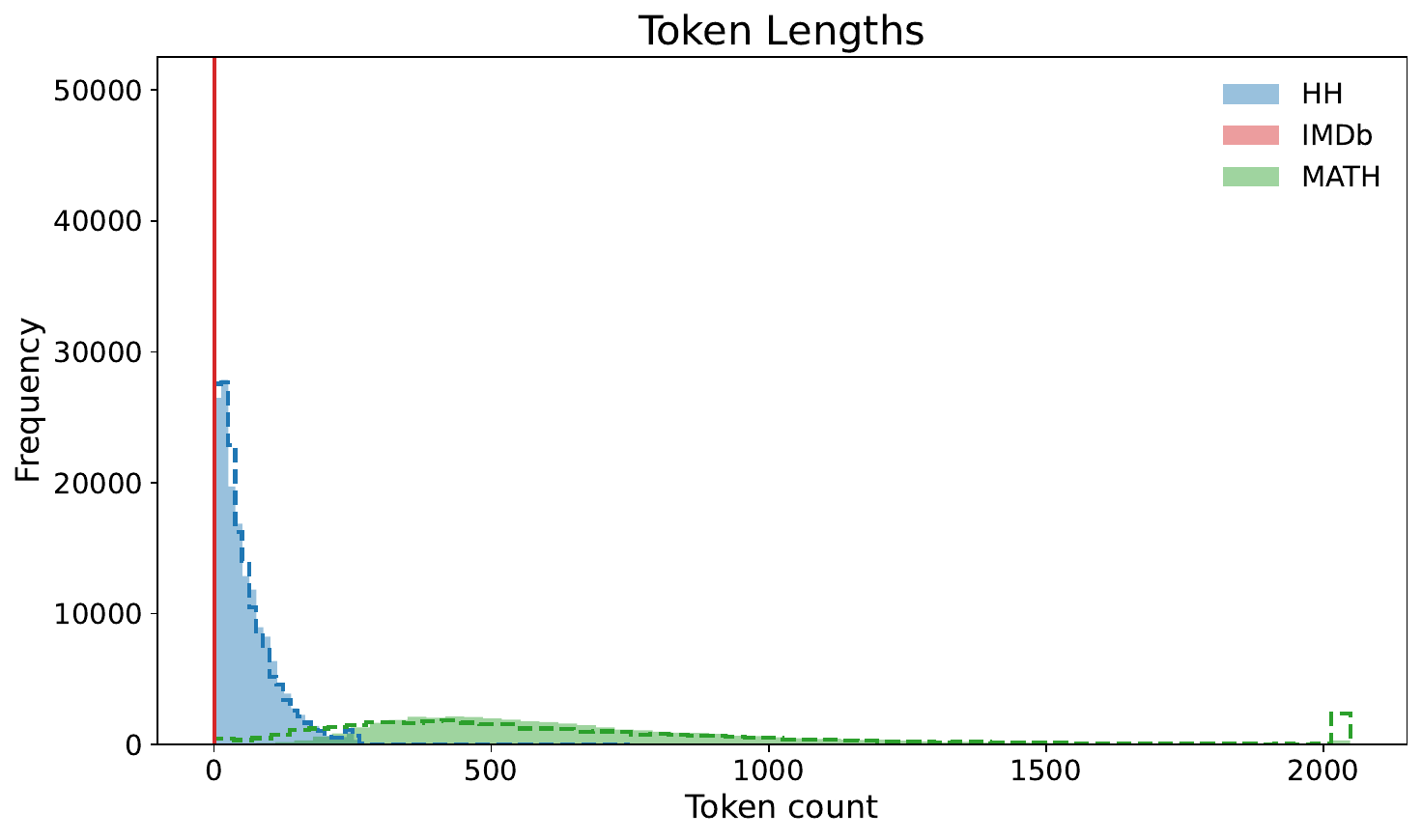}
    \caption{Token lengths for each dataset.  IMDb consists of 1 token long trajectories, HH with moderately long trajectories, while MATH consists of long trajectories.}
    \label{fig:token_len}
\end{figure}

This section provides the detailed settings for the pilot experiment in Figure~\ref{fig:grad_ent}.
\paragraph{Datasets.} To test the gradient entanglement between preferred and dispreferred trajectories, we plot their angles and norms for three different datasets.  The difference in trajectory length between the three datasets is illustrated in Figure~\ref{fig:token_len}.
\begin{itemize}
    \item \textbf{Stanford IMDb} \citep{maas2011learning}: A movie classification dataset with single-token (Positive/Negative) preferred and dispreferred trajectories.
    \item \textbf{Anthropic Helpful and Harmless} \citep{bai2022training}: A commonly used RLHF dataset with moderately long preferred and dispreferred trajectories.
    \item \textbf{Hendrycks MATH dataset} \citep{hendrycks2021math}: A commonly used math dataset with long, gradient entangled preferred and dispreferred trajectories.
\end{itemize}

\paragraph{Training Settings.}
We fine-tune the Qwen2.5-Math-1.5B model \citep{yang2024qwenmath} on the IMDb, HH, and MATH datasets. For the MATH dataset, we use the same preference data as described below for the Qwen2.5-Math-7B model. For IMDb and HH, we use the originally provided datasets. We conduct all training on a single NVIDIA A100 GPU.  We use a cosine learning rate scheduler.  Table~\ref{tab:pilot_params} shows the detailed hyperparameters. 
\begin{table}[!ht]
\centering
\caption{Pilot experiment hyperparameters.}
\label{tab:pilot_params}
\begin{tabular}{ll}
\toprule
\textbf{Parameter} & \textbf{Value} \\
\midrule
Seed & 42 \\
Optimizer & AdamW ($\beta_1=0.9, \beta_2=0.999$) \\
Warmup Steps & 150 \\
Learning Rate & $5\times10^{-6}$ \\
Max Gradient Norm & 10 \\
Max Steps & 25{,}000 \\
Batch Size & 64 \\
Precision & \texttt{bfloat16} \\
Max Sequence Length & 2{,}048 \\
\bottomrule
\end{tabular}
\end{table}

\subsection{Main Dataset Construction}
For our main experiments, we employ the 12K training set of the MATH dataset following \citet{lightman2024let}, and sample $K=8$ trajectories per problem. We use regex matching to identify correct answers. Correct trajectories are rewarded $+1$ 
while incorrect trajectories are rewarded $-1$. Following \citet{yu2025dapo}, 
any questions where generated answers are all correct or all are incorrect are 
discarded. Temperature sampling ($T=0.7$) is used for dataset generation with 
a constant seed of 42, and 5\% of the data is held out for validation. For code generation, we use the same construction procedure, but with TACO~\citep{li2023taco} and APPS~\citep{hendrycks2021apps} datasets for training. All inference is performed on a single node with two NVIDIA H200 GPUs with vLLM \citep{kwon2023efficient}.
\subsection{Algorithms}
This section details the baseline methods used in our experiments.
\begin{itemize}[leftmargin=*]
\item\textbf{Supervised Fine-Tuning (SFT)} or Rejection Fine-Tuning (RFT) \citep{yuan2023rft} fine-tunes the model using standard negative log-likelihood loss, exclusively on the correct trajectories ($y_w$). The loss function is:
\begin{align*}
    \mathcal{L}_{\text{SFT}} = - \mathbb{E}_{(x, y_w) \sim \mathcal{D}} [\log \pi_\theta(y_w \mid x)]
\end{align*}

\item\textbf{Direct Preference Optimization (DPO)}
\citep{rafailov2024dpo} solves the closed-form solution of KL-regularized RL with the Bradley-Terry assumption \citep{Bradley1952}. DPO is now one of the most popular offline RL algorithms in domains like preference alignment. The loss function is:
\begin{align*}
    \mathcal{L}_{\text{DPO}}(\pi_\theta; \pi_{\text{ref}}) = -\mathbb{E}_{(x, y_w, y_l) \sim \mathcal{D}}
    \left[\log \sigma\!\left(
        \beta \log \frac{\pi_\theta(y_w \mid x)}{\pi_{\text{ref}}(y_w \mid x)}
        - \beta \log \frac{\pi_\theta(y_l \mid x)}{\pi_{\text{ref}}(y_l \mid x)}
    \right)\right]
\end{align*}
where $\sigma$ is the sigmoid function, $\beta$ controls the deviation from the reference policy $\pi_{\text{ref}}$, and $y_l$ denotes the incorrect response. We set $\beta=0.1$ throughout our work following \citet{pang2024iterativerpo, jiao2024learning, tu2025dpo}.

\item\textbf{Reasoning Preference Optimization (RPO)} \citep{pang2024iterativerpo} adds a length-normalized negative log-likelihood term to DPO to mitigate the decrease of probabilities in rewarded samples in reasoning domains. RPO shows promising performance compared to baseline DPO in reasoning domains. The loss function is:
\begin{align*}
    \mathcal{L}_{\text{RPO}} = \mathcal{L}_{\text{DPO}}(\pi_\theta; \pi_{\text{ref}})
    - \alpha\, \mathbb{E}_{(x, y_w) \sim \mathcal{D}} [\frac{1}{|y_w|}\log \pi_\theta(y_w \mid x)]
\end{align*}
where $\alpha > 0$ controls the weight of the NLL regularization on correct responses, following the original paper we set $\alpha = 1$.

\item\textbf{DPO-Positive (DPOP)} \citep{pal2024smaug} augments DPO with a penalty that prevents the preferred response likelihood from decreasing excessively. We tune its regularization strength $\lambda$ and use $\lambda=50$ in the main Qwen3-4B-Base comparison. The loss function is:
\begin{align*}
    \mathcal{L}_{\text{DPOP}} = -\,\mathbb{E}_{(x, y_w, y_l) \sim \mathcal{D}} \Bigg[ \log \sigma \Bigg(
    &\beta \log \frac{\pi_\theta(y_w \mid x)}{\pi_{\text{ref}}(y_w \mid x)}
    - \beta \log \frac{\pi_\theta(y_l \mid x)}{\pi_{\text{ref}}(y_l \mid x)} - \lambda_\text{DPOP} \cdot \max\Big(0,\ \log \frac{\pi_{\text{ref}}(y_w \mid x)}{\pi_\theta(y_w \mid x)}\Big)
    \Bigg) \Bigg]
\end{align*}
where $\lambda_\text{DPOP} > 0$ controls the strength of the penalty, which activates only when the policy assigns lower likelihood to the preferred response $y_w$ than the reference model does.

\item\textbf{Kahneman-Tversky Optimization (KTO)} \citep{Ethayarajh2024KTO} models alignment based on Kahneman-Tversky's Prospect Theory \citep{kahneman1979prospect}. KTO is a widely used alignment method that does not require paired preference data. The KTO loss is defined as:
\begin{align*}
    \mathcal{L}_{\text{KTO}}(\pi_\theta, \pi_{\text{ref}}) &= \mathbb{E}_{(x,y) \sim \mathcal{D}}[\lambda_y - v(x,y)] \\
    r_\theta(x,y) &= \log \frac{\pi_\theta(y \mid x)}{\pi_{\text{ref}}(y\mid x)} \\
    z_0 &= \frac{1}{|\mathcal{D}|}\sum_{(x',y') \in \mathcal{D}} r_\theta(x',y') \\
    v(x,y) &= \begin{cases}
    \lambda_w \sigma(\beta(r_\theta(x,y) - z_0)) & \text{if } y \text{ is correct} \\
    \lambda_l \sigma(\beta(z_0 - r_\theta(x,y))) & \text{if } y \text{ is incorrect}
    \end{cases}
\end{align*}
Here, $\lambda_y$ is either $\lambda_w$ or $\lambda_l$ depending on whether $y$ is correct or incorrect, and $z_0$ represents the mean log-likelihood ratio over the dataset. Following the original paper, we set $\lambda_w=\lambda_l=1$ by default.

\item\textbf{Policy Optimization via Optimal Advantage Regression (A*-PO)} 
\citep{brantley2025accelerating} solves the KL-regularized RL objective through 
two-stage regression and does not rely on the Bradley-Terry assumption. The first 
stage approximates the optimal value $(\hat{V}^\star(x))$ with Monte Carlo estimates.
\begin{align*}
    \hat{V}^\star(x) \leftarrow \beta_1 \ln \left( \frac{1}{N} \sum_{i=1}^{N} 
    \exp\!\left(\frac{r(x, y_i)}{\beta_1}\right)\right), \quad y_i \sim 
    \pi_{\mathrm{ref}}(\cdot \mid x),
\end{align*}
Then, the second stage optimizes the closed form of KL-regularized RL using 
$\hat{V}^\star(x)$ inferred in the first stage. While the original A*-PO proposed 
an online second iteration, for a fair comparison we consider an offline version 
of the second stage. The loss function is:
\begin{align*}
    \mathcal{L}_{A^\star\text{-PO}}(\pi; \pi_t) =
    \mathbb{E}_{x,\, y \sim \pi_t(\cdot|x)}
    \left[
        \left(
            \beta_2 \log \frac{\pi(y\mid x)}{\pi_{\text{ref}}(y\mid x)}
            - \bigl(r(x,y) - \hat{V}^\star(x)\bigr)
        \right)^2
    \right]
\end{align*}
where $r(x,y) - \hat{V}^\star(x)$ serves as an estimate of the optimal advantage $A^\star(x,y)$.  Following the original paper, we set $\beta_1=0.5$ and $\beta_2 = 1\times10^{-3}$.
\end{itemize}

\subsection{Evaluation Protocol}
\paragraph{Benchmarks}
The following section details the benchmarks used in our experiments and our evaluation protocol.
\begin{itemize}[leftmargin=*]
  \item \textbf{GSM8K} \citep{cobbe2021gsm8k} is a dataset of 8,500 grade-school-level math word problems (1,319 test problems) requiring multi-step arithmetic reasoning. It is widely used as a standard benchmark for elementary mathematical reasoning ability. We evaluate on the full 1,319 test samples.

  \item \textbf{MATH500}~\citep{hendrycks2021math} is a widely used, representative 500-problem subset of the MATH benchmark test set, spanning seven mathematical subjects (algebra, geometry, number theory, etc.) at difficulty levels 1 to 5.

  \item \textbf{MATH-P} \citep{huang2025mathp} consists of perturbed variants of MATH benchmark
  problems created by modifying numerical values and conditions. It includes 279 questions each for MATH-P Simple, which applies minor numerical perturbations, and MATH-P Hard, which applies more substantial structural perturbations.

  \item \textbf{GaoKao2023En} \citep{liao2024gaokao} is the English-translated version of the 2023 Chinese College Entrance
  Examination (Gaokao) mathematics problems. We adopt the data from Qwen2.5-Math \citep{yang2024qwenmath} evaluation, consisting of 385 problems.

  \item \textbf{OlympiadBench} \citep{he2024olympiadbench} is a benchmark comprising
  competition-level mathematics and physics problems drawn from Olympiad competitions. Following the Qwen2.5-Math~\citep{yang2024qwenmath} evaluation, we adopt the English mathematics subset, consisting of 675 problems.

  \item \textbf{AMC}, or the American Mathematics Competition, is a series of multiple-choice
  mathematics contests for high school students that test problem-solving ability across a range of
  topics. We employ the recent 2023 and 2024 AMC datasets, consisting of 90 problems in total (45 per year).

  \item \textbf{AIME}, or the American Invitational Mathematics Examination, is a prestigious competition featuring challenging problems that require sophisticated mathematical reasoning. We employ the recent 2024, 2025, and 2026 datasets, consisting of 90 problems in total (30 per year).

  \item \textbf{MBPP+}~\citep{liu2023your} is the strengthened version of the Mostly Basic Python Problems (MBPP) benchmark introduced by EvalPlus. It contains 378 Python programming problems and substantially expands the original unit-test coverage, providing a more rigorous evaluation of functional correctness.
  
  \item \textbf{BigCodeBench (BCB)}~\citep{zhuo2025bigcodebench} evaluates practical code generation requiring complex instructions and diverse function calls. It contains 1,140 programming tasks covering 723 function calls from 139 libraries across seven application domains, with high-coverage execution tests for each task. We employ the calibrated BCB for evaluation.
  
  \item \textbf{LiveCodeBench v6 (LCB)}~\citep{jain2025livecodebench} is a contamination-resistant code-generation benchmark constructed from recently released competitive-programming problems. We use the code-generation v6 evaluation split, consisting of 175 problems spanning easy, medium, and hard difficulty levels.
\end{itemize}
\paragraph{Evaluation Protocol.}
We evaluate model performance on the basis of Pass@1 accuracy on the respective test sets. All inference is conducted using the vLLM library \citep{kwon2023efficient} on a single node with two NVIDIA H200 GPUs. For each problem, we generate $n=8$ independent completions (and $n=32$ for AMC and AIME) at temperature 0.7. Let $z_{ij}\in\{0,1\}$ denote whether the $j$-th completion for problem $i$ is correct, and let $a_j=\frac{1}{N}\sum_{i=1}^{N} z_{ij}$
denote the Pass@1 obtained from the $j$-th independent sample across the benchmark. We report the mean repeated-sampling estimate:
\begin{equation}
    \widehat{\text{Pass@1}}=\frac{1}{n}\sum_{j=1}^{n} a_j =\frac{1}{N}\sum_{i=1}^{N}\frac{1}{n}\sum_{j=1}^{n} z_{ij}.
\end{equation}
This estimates the expected single-sample accuracy under stochastic decoding.

\subsection{Training Details}
We conduct all training on a single node with two NVIDIA H200 GPUs. Since the learning rate plays an important role in post-training, we conduct a comprehensive search. We use a learning rate of $5\times10^{-6}$ for Qwen3-4B-Base, Qwen2.5-Math-7B, and $3\times10^{-6}$ for Llama-3.2-3B-Instruct. The Supervised Fine-Tuning (SFT) phase uses a standard, higher learning rate of $2\times10^{-5}$. We employ early stopping based on the performance of the validation set. Table~\ref{tab:general_params} further details hyperparameters.

\paragraph{Computational Overhead of \method{}.}
Table~\ref{tab:fpa_efficiency} compares the training efficiency of
\method{} and Off-RL under the same hardware (NVIDIA H200) and training configuration. \method{} decreases throughput by approximately 3\%, from 16.6 to 16.1 examples/s, and increases peak GPU memory from 96 to 105 GB

\begin{table}[!ht]
\centering
\caption{\textbf{Computational overhead of \method{}.}
Throughput and peak GPU memory measured under the same hardware and
training configuration.}
\label{tab:fpa_efficiency}
\small
\setlength{\tabcolsep}{8pt}
\begin{tabular}{lcc}
\toprule
\textbf{Method}
& \textbf{Throughput (examples/s) $\uparrow$}
& \textbf{Peak GPU memory (GB) $\downarrow$} \\
\midrule
Off-RL                & 16.6 & 96  \\
Off-RL w/ \method{}   & 16.1 & 105 \\
\bottomrule
\end{tabular}
\end{table}

\subsection{Prompts}
We use the following prompt in Figure~\ref{fig:baseprompt} for both training and evaluation in our main mathematical experiments.

\begin{figure*}[!ht]
\centering
\caption{\textbf{Prompt} used for mathematical experiments.}
\fbox{%
\begin{minipage}{0.94\textwidth}
\small\ttfamily
Solve this math problem step by step. At the end, make sure to finish the calculation and state the answer exactly once in the following format:
The final answer is \textbackslash boxed\{X\}, where X is your final answer.

\medskip
Q: \{Question\}

A:
\end{minipage}%
}
\label{fig:baseprompt}
\end{figure*}

\begin{table}[!ht]
\centering
\caption{\textbf{Hyperparameters} used for the main experiment.}
\small
\setlength{\tabcolsep}{2pt}
\renewcommand{\arraystretch}{1.15}
\begin{tabular}{@{}l r@{}}
\toprule
\textbf{Parameter} & \textbf{Value} \\
\midrule
Seed & 42 \\
Optimizer & AdamW ($\beta_1{=}0.9$, $\beta_2{=}0.999$) \\
Warmup Steps & 150 \\
Learning Rate & $\{5, 3, 1, 0.5\}\times10^{-6}$ \\
Max Gradient Norm & 10 \\
Max Epochs & 3 \\
Batch Size & 64 \\
Precision & \texttt{bfloat16} \\
Max Generation Length & 2{,}048 \\
\bottomrule
\end{tabular}
\label{tab:general_params}
\end{table}

\subsection{KL-Regularized Off-RL}
\label{app:kl_ablation}
This section details the KL-regularization discussed in \S~\ref{sec:exp_further}. The KL divergence between two distributions $p$ and $q$ is asymmetric ($\mathbb{D}_{\text{KL}}(p \| q) \neq \mathbb{D}_{\text{KL}}(q \| p)$). Each direction induces different behaviors and requires a different sampling distribution.
\textbf{Forward KL} penalizes $\pi_\theta$ for assigning low probability to regions where the reference $\pi_{\text{ref}}$ places high mass:
\begin{equation}
    \mathbb{D}_{\text{KL}}(\pi_{\text{ref}} \| \pi_\theta) = \mathbb{E}_{y \sim \pi_{\text{ref}}}\!\left[\log \frac{\pi_{\text{ref}}(y \mid x)}{\pi_\theta(y \mid x)}\right].
\end{equation}
Crucially, computing an unbiased gradient estimate of the forward KL requires samples from $\pi_{\text{ref}}$ rather than $\pi_\theta$.
\textbf{Reverse KL} penalizes the policy $\pi_\theta$ for assigning high probability to regions where the reference $\pi_{\text{ref}}$ places low:
\begin{equation}
    \mathbb{D}_{\text{KL}}(\pi_\theta \| \pi_{\text{ref}}) = \mathbb{E}_{y \sim \pi_\theta}\!\left[\log \frac{\pi_\theta(y \mid x)}{\pi_{\text{ref}}(y \mid x)}\right].
\end{equation}
This term is tractable when sampling with respect to $\pi_\theta$, which is the standard paradigm in online learning, thus is frequently used in online RL.
\paragraph{Why Forward KL in the Offline Setting.}
In our offline setting, the training dataset $\mathcal{D}$ is generated once from a fixed reference policy $\pi_{\text{ref}}$ and reused across training. Because it is computable in this context, we utilize the forward KL. We therefore augment Off-RL with a forward KL penalty:
\begin{equation}
  \mathcal{J}_{\text{KL}}(\theta) = \mathcal{J}(\theta) - \tau\, \mathbb{E}_{x \sim \mathcal{D}}\!\left[\mathbb{D}_{\text{KL}}(\pi_\text{ref}(\cdot \mid x) \| \pi_{\theta}(\cdot \mid x))\right].  
\end{equation}

\subsection{RLVR Training}
\label{app:dapo}
This section details the RLVR training settings introduced in \S~\ref{sec:exp_grpo}. We initialize our policy model with Qwen3-4B-Base and train using the same MATH train dataset. The training is performed on a single node with 2 NVIDIA H200 GPUs for a fair comparison, and a complete set of hyperparameters is in Table~\ref{tab:rlvr_params}.

\paragraph{GRPO.}
We employ Group Relative Policy Optimization (GRPO)~\citep{shao2024deepseekmath} as a representative critic-free RLVR algorithm. For each prompt $x$, GRPO samples a group of $G=8$ responses $\{y_1,\dots,y_G\}$ from the old policy $\pi_{\theta_{\mathrm{old}}}$. Rather than learning a separate value function, GRPO uses the rewards of responses within the same group to construct a relative baseline. Under our trajectory-level verifiable reward setting, every token in response $y_i$ shares the same group-normalized advantage:
\begin{equation}
    \hat{A}_{i,t}
    =
    \frac{
        r_i - \mathrm{mean}(\{r_j\}_{j=1}^{G})
    }{
        \mathrm{std}(\{r_j\}_{j=1}^{G})
    }.
\end{equation}
GRPO then optimizes a PPO-style clipped surrogate objective. Defining the token-level importance ratio as
\begin{equation}
    \rho_{i,t}
    =
    \frac{
        \pi_\theta(y_{i,t}\mid x,y_{i,<t})
    }{
        \pi_{\theta_{\mathrm{old}}}(y_{i,t}\mid x,y_{i,<t})
    },
\end{equation}
the GRPO objective is
\begin{equation}
    \mathcal{L}_{\mathrm{GRPO}}(\theta)
    =
    -\mathbb{E}_{x,\{y_i\}}
    \left[
        \frac{1}{G}
        \sum_{i=1}^{G}
        \frac{1}{|y_i|}
        \sum_{t=1}^{|y_i|}
        \min\left(
            \rho_{i,t}\hat{A}_{i,t},
            \mathrm{clip}(\rho_{i,t},1-\epsilon,1+\epsilon)\hat{A}_{i,t}
        \right)
    \right].
\end{equation}

\paragraph{DAPO.}
We additionally employ Decoupled Clip and Dynamic Sampling Policy Optimization (DAPO)~\citep{yu2025dapo}, which improves upon GRPO with modifications designed to stabilize and improve the efficiency of long chain-of-thought RLVR training. 
\begin{itemize}
    \item \textit{Clip-Higher} replaces the symmetric clipping range of GRPO with separate lower and upper bounds, $\epsilon_{\mathrm{low}}$ and $\epsilon_{\mathrm{high}}$, where $\epsilon_{\mathrm{high}}>\epsilon_{\mathrm{low}}$. This provides more room for low-probability tokens to be promoted and helps mitigate entropy collapse.
    \item \textit{Dynamic Sampling} removes prompts for which all $G$ responses are correct or all are incorrect. Because such groups have identical rewards, their group-normalized advantages vanish and they provide no useful gradient signal; DAPO therefore continues sampling until the target batch contains informative groups with both correct and incorrect responses.
    \item DAPO also aggregates the policy-gradient loss at the token level rather than averaging each response equally. Following the original DAPO setting, we also omit the KL penalty to the reference policy.
\end{itemize}

The resulting objective is
\begin{equation}
\begin{aligned}
    \mathcal{L}_{\mathrm{DAPO}}(\theta)
    =
    -\mathbb{E}_{x,\{y_i\}}
    \Bigg[
        \frac{1}{\sum_{i=1}^{G}|y_i|}
        \sum_{i=1}^{G}
        \sum_{t=1}^{|y_i|}
        \min\Big(
            &\rho_{i,t}\hat{A}_{i,t},\mathrm{clip}(
                \rho_{i,t},
                1-\epsilon_{\mathrm{low}},
                1+\epsilon_{\mathrm{high}}
            )\hat{A}_{i,t}
        \Big)
    \Bigg],
\end{aligned}
\end{equation}
where dynamic sampling retains only groups satisfying
\begin{equation}
    0
    <
    \left|
        \{y_i : \mathrm{is\_correct}(y_i)\}
    \right|
    <
    G.
\end{equation}

\begin{table}[!ht]
\centering
\caption{\textbf{Hyperparameters} used for GRPO and DAPO training.}
\small
\setlength{\tabcolsep}{2pt}
\renewcommand{\arraystretch}{1.15}
\begin{tabular}{@{}l r@{}}
\toprule
\textbf{Parameter} & \textbf{Value} \\
\midrule
Seed & 42 \\
Training Steps & 1{,}500 \\
Learning Rate & $1 \times 10^{-6}$ \\
Global Batch Size & 128 \\
Mini-batch Size & 64 \\
Micro-batch per GPU & 8 \\
Group Size ($G$) & 8 \\
Temperature & 0.7 \\
Max Generation Length & 2{,}048 \\
GRPO Clip $\epsilon$ & 0.2 \\
DAPO Clip $\epsilon_{\mathrm{low}},\epsilon_{\mathrm{high}}$ & 0.2, 0.28 \\
\bottomrule
\end{tabular}
\label{tab:rlvr_params}
\end{table}

\section{Additional Results}
\subsection{Hyperparameter Selection and Ablation}
\label{app:hyperparameter}

We select the extrapolation strength $\lambda$ using a held-out validation set. We search over $\lambda \in \{0.5, 1, 2, 5, 10\}$ for all models and additionally include $\lambda=20$ for Qwen3-4B-Base. As shown in Table~\ref{tab:lambda_ablation_val}, we select $\lambda=2$ for Qwen3-4B-Base and $\lambda=1$ for both Qwen2.5-Math-7B and Llama-3.2-3B-Instruct. These selected values are used throughout the main experiments.

We report the full test-set ablation of $\lambda$ in Table~\ref{tab:lambda_ablation}. Overall, $\lambda \in \{1,2\}$ performs best across the three model families. Larger values can reduce the gains, but generally remain competitive with Off-RL ($\lambda=0$), and remain above it for Qwen3-4B-Base. The broad range of competitive values indicates that \method{} is reasonably robust to the choice of $\lambda$, with values around 1 (e.g., $0.5$, $1$, and $2$) generally performing well across models. 

\begin{table*}[!tb]
\centering
\caption{\textbf{Validation set} results of the hyperparameter $\lambda$ across models. Our selected hyperparameters are highlighted in gray. All results are reported as Pass@1 accuracy. The best performance is \textbf{bolded}.}
\small
\setlength{\tabcolsep}{6pt}
\renewcommand{\arraystretch}{1.15}
\begin{tabularx}{\textwidth}{l *{6}{C}}
\toprule
\textbf{Model} & $\lambda=0.5$ & $\lambda=1$ & $\lambda=2$ & $\lambda=5$ & $\lambda=10$ & $\lambda=20$ \\
\midrule
Qwen3-4B-Base         & 74.70 & 79.65 & \cellcolor{rowhl}\textbf{83.00} & 76.26 & 76.64 & 75.30 \\
Qwen2.5-Math-7B       & 74.48 & \cellcolor{rowhl}\textbf{75.56} & 74.39 & 74.59 & 74.66 & $-$ \\
Llama-3.2-3B-Instruct & 65.65 & \cellcolor{rowhl}\textbf{66.34} & 65.39 & 64.05 & 64.87 & $-$ \\
\bottomrule
\end{tabularx}
\label{tab:lambda_ablation_val}
\end{table*}

\begin{table*}[!t]
\centering
\caption{\textbf{Effect of $\lambda$} across multiple mathematical reasoning datasets and models. Our selected hyperparameters are highlighted. All results are reported as Pass@1 accuracy. The best performance is \textbf{bolded}, and the second best is \underline{underlined}.}
\small
\setlength{\tabcolsep}{3pt}
\renewcommand{\arraystretch}{1.15}
\begin{tabularx}{\textwidth}{l *{8}{C}}
\toprule
\textbf{Method} & \textbf{GSM8K} & \textbf{MATH500} & \textbf{MATH-P} & \textbf{GaoKao} & \textbf{Olym.} & \textbf{AMC} & \textbf{AIME} & \textbf{Avg.} \\
\midrule
\grouphead{9}{Qwen3-4B-Base} \\
Base      & 75.52 & 56.83 & 33.18 & 43.96 & 25.40 & 29.78 & 6.60 & 38.75 \\
$\lambda = 0$   & 86.80 & 66.70 & 41.42 & 52.76 & 30.72 & 36.91 & 8.50 & 46.26 \\
$\lambda = 0.5$ & 83.86 & 67.20 & 43.53 & 52.05 & 31.54 & 37.83 & 7.08 & 46.16 \\
$\lambda = 1$   & \underline{88.69} & \underline{70.90} & \underline{44.76} & \underline{55.16} & \underline{33.91} & \underline{37.92} & 9.72 & \underline{48.72} \\
\rowcolor{rowhl}
$\lambda = 2$   & \textbf{91.60} & \textbf{74.50} & \textbf{48.40} & \textbf{59.60} & \textbf{35.40} & \textbf{42.80} & \underline{10.10} & \textbf{51.77} \\
$\lambda = 5$   & 86.64 & 68.45 & 43.26 & 55.10 & 33.02 & 37.28 & 8.06 & 47.40 \\
$\lambda = 10$  & 84.68 & 67.45 & 43.55 & 54.87 & 31.57 & 37.14 & \textbf{11.25} & 47.22 \\
$\lambda = 20$  & 83.90 & 67.67 & 43.84 & 55.03 & 32.78 & 37.19 & 9.72 & 47.16 \\
\midrule
\grouphead{9}{Qwen2.5-Math-7B} \\
Base      & 77.17 & 63.65 & 39.79 & 47.73 & 28.96 & 36.79 & 10.80 & 43.56 \\
$\lambda = 0$   & 80.39 & 68.73 & \underline{43.91} & 53.05 & 31.72 & 41.99 & 11.84 & 47.38 \\
$\lambda = 0.5$ & 81.00 & 68.45 & 43.67 & 53.18 & 33.11 & \textbf{43.87} & \underline{12.01} & 47.90 \\
\rowcolor{rowhl}
$\lambda = 1$   & \textbf{83.95} & \textbf{72.55} & \textbf{44.58} & \textbf{56.07} & \textbf{33.50} & 42.32 & \textbf{12.74} & \textbf{49.39} \\
$\lambda = 2$   & 82.39 & \underline{70.17} & 42.97 & 54.64 & \underline{33.37} & \underline{43.66} & 11.67 & \underline{48.41} \\
$\lambda = 5$   & \underline{83.69} & 69.55 & 40.35 & \underline{55.19} & 32.00 & 42.03 & 11.11 & 47.70 \\
$\lambda = 10$  & 77.00 & 66.97 & 39.96 & 49.00 & 32.54 & 40.84 & 11.91 & 45.46 \\
\midrule
\grouphead{9}{Llama-3.2-3B-Instruct} \\
Base      & 77.52 & 43.20 & 20.88 & 32.99 & 13.80 & 16.34 & \underline{2.71} & 29.63 \\
$\lambda = 0$   & 78.75 & 43.75 & 21.13 & 33.90 & 13.72 & 14.14 & 2.29 & 29.67 \\
$\lambda = 0.5$ & 80.22 & 44.73 & \textbf{21.94} & \textbf{34.29} & 13.87 & 17.46 & 2.64 & 30.74 \\
\rowcolor{rowhl}
$\lambda = 1$   & \textbf{80.54} & \textbf{46.02} & 20.70 & \underline{34.16} & \underline{14.13} & \underline{18.37} & 2.08 & \underline{30.86} \\
$\lambda = 2$   & \underline{80.31} & \underline{45.85} & \underline{21.20} & 33.64 & \textbf{14.31} & \textbf{19.05} & 1.95 & \textbf{30.90} \\
$\lambda = 5$   & 79.29 & 43.58 & 18.86 & 31.98 & 12.37 & 16.24 & 2.64 & 29.28 \\
$\lambda = 10$  & 79.22 & 43.50 & 18.55 & 31.66 & 12.30 & 16.79 & \textbf{3.41} & 29.35 \\
\bottomrule
\end{tabularx}
\label{tab:lambda_ablation}
\end{table*}

\subsection{Baseline Hyperparameter Tuning}
\label{app:baseline_tuning}

To ensure that our comparisons are not due to under-tuned baselines, we sweep the principal hyperparameters of each method on Qwen3-4B-Base and report the full results in Table~\ref{tab:baseline_tuning}. 
For each method, we vary its main optimization or method-specific hyperparameter, including the learning rate and the corresponding regularization or loss coefficient. 
Settings marked $\dagger$ are used in the main experiments. 
Even against the best tuned configuration of each baseline, \method{} retains a clear margin: the strongest baseline reaches an average accuracy of $48.41$, compared with $51.77$ for \method{}.

\begin{table*}[!t]
\centering
\caption{\textbf{Hyperparameter tuning} on Qwen3-4B-Base. $\dagger$ marks the configuration used in the main tables.}
\small
\setlength{\tabcolsep}{2pt}
\renewcommand{\arraystretch}{1.15}
\begin{tabularx}{\textwidth}{l *{8}{C}}
\toprule
\textbf{Method} & \textbf{GSM8K} & \textbf{MATH500} & \textbf{MATH-P} & \textbf{GaoKao} & \textbf{Olym.} & \textbf{AMC} & \textbf{AIME} & \textbf{Avg.} \\
\midrule
\grouphead{9}{DPO --- learning rate $\eta$} \\
\hspace{0.4em}$5\times10^{-7}$ & 75.62 & 56.57 & 33.54 & 43.31 & 25.98 & 30.63 & 6.84 & 38.93 \\
\hspace{0.4em}$1\times10^{-6}$ & 75.66 & 58.48 & 33.63 & 43.93 & 26.19 & 29.60 & 6.77 & 39.18 \\
\hspace{0.4em}$5\times10^{-6}\,^{\dagger}$ & 85.05 & 69.35 & 43.66 & 55.13 & 32.76 & 39.39 & 9.58 & 47.85 \\
\hspace{0.4em}$1\times10^{-5}$ & 67.64 & 54.83 & 25.88 & 45.39 & 20.56 & 23.42 & 4.72 & 34.63 \\
\midrule
\grouphead{9}{DPO --- KL strength $\beta$} \\
\hspace{0.4em}$0.05$ & 82.91 & 69.58 & 44.06 & 54.06 & 33.19 & 38.08 & 9.20 & 47.30 \\
\hspace{0.4em}$0.1\,^{\dagger}$ & 85.05 & 69.35 & 43.66 & 55.13 & 32.76 & 39.39 & 9.58 & 47.85 \\
\hspace{0.4em}$0.2$ & 83.12 & 65.90 & 41.92 & 52.73 & 30.30 & 35.59 & 9.03 & 45.51 \\
\hspace{0.4em}$0.3$ & 82.70 & 64.35 & 40.39 & 51.53 & 30.11 & 35.60 & 8.68 & 44.77 \\
\midrule
\grouphead{9}{RPO --- NLL weight $\alpha$} \\
\hspace{0.4em}$0.25$ & 82.05 & 65.62 & 40.73 & 51.79 & 30.52 & 34.48 & 9.06 & 44.89 \\
\hspace{0.4em}$0.5$ & 81.98 & 64.40 & 39.65 & 51.23 & 29.72 & 34.94 & 8.57 & 44.36 \\
\hspace{0.4em}$1\,^{\dagger}$ & 88.61 & 64.70 & 38.98 & 52.37 & 30.37 & 33.54 & 8.50 & 45.30 \\
\hspace{0.4em}$2$ & 81.37 & 62.28 & 37.99 & 48.21 & 28.61 & 32.91 & 8.06 & 42.78 \\
\midrule
\grouphead{9}{DPOP --- regularization strength $\lambda_\text{DPOP}$} \\
\hspace{0.4em}$5$ & 82.43 & 64.90 & 41.20 & 50.26 & 29.94 & 35.30 & 8.06 & 44.58 \\
\hspace{0.4em}$50\,^{\dagger}$ & 82.30 & 64.88 & 41.46 & 51.36 & 30.06 & 35.60 & 8.92 & 44.94 \\
\hspace{0.4em}$500$ & 83.06 & 65.00 & 40.48 & 50.81 & 30.15 & 35.40 & 8.65 & 44.79 \\
\midrule
\grouphead{9}{KTO --- $\beta$} \\
\hspace{0.4em}$0.05\,^{\dagger}$ & 86.50 & 67.60 & 40.24 & 53.44 & 31.44 & 36.40 & 7.99 & 46.23 \\
\hspace{0.4em}$0.1$ & 81.71 & 66.88 & 41.07 & 52.21 & 31.78 & 35.37 & 8.85 & 45.41 \\
\hspace{0.4em}$0.3$ & 85.26 & 66.83 & 40.66 & 51.79 & 30.57 & 35.11 & 8.27 & 45.50 \\
\midrule
\grouphead{9}{KTO --- $\lambda_w=\lambda_l$} \\
\hspace{0.4em}$0.5$ & 83.06 & 60.25 & 35.89 & 49.09 & 26.04 & 29.25 & 6.01 & 41.37 \\
\hspace{0.4em}$1\,^{\dagger}$ & 86.50 & 67.60 & 40.24 & 53.44 & 31.44 & 36.40 & 7.99 & 46.23 \\
\hspace{0.4em}$2$ & 76.47 & 65.88 & 41.58 & 54.45 & 32.00 & 37.20 & 9.03 & 45.23 \\
\midrule
\grouphead{9}{A*-PO --- $\beta_2$} \\
\hspace{0.4em}$1\times10^{-4}$ & 70.86 & 50.42 & 28.41 & 36.10 & 21.76 & 21.58 & 6.94 & 33.72 \\
\hspace{0.4em}$1\times10^{-3}\,^{\dagger}$ & 86.65 & 63.82 & 42.57 & 48.12 & 31.40 & 33.30 & 7.88 & 44.82 \\
\midrule
\grouphead{9}{Off-RL --- learning rate $\eta$} \\
\hspace{0.4em}$5\times10^{-7}$ & 75.59 & 58.65 & 34.84 & 44.03 & 26.59 & 29.77 & 7.29 & 39.54 \\
\hspace{0.4em}$1\times10^{-6}$ & 80.30 & 64.03 & 39.90 & 50.36 & 29.87 & 33.99 & 8.13 & 43.80 \\
\hspace{0.4em}$5\times10^{-6}\,^{\dagger}$ & 86.80 & 66.70 & 41.42 & 52.76 & 30.72 & 36.91 & 8.50 & 46.26 \\
\hspace{0.4em}$1\times10^{-5}$ & 67.64 & 52.28 & 31.21 & 43.15 & 23.09 & 33.99 & 6.11 & 36.78 \\
\midrule
\grouphead{9}{Off-RL w/ RS --- incorrect-gradient scale} \\
\hspace{0.4em}$0.3$ & 86.10 & 65.00 & 38.10 & 52.40 & 30.40 & 30.80 & 7.10 & 44.27 \\
\hspace{0.4em}$0.5$ & 86.90 & 65.90 & 39.90 & 53.80 & 30.30 & 30.60 & 8.20 & 45.09 \\
\midrule
\grouphead{9}{Off-RL w/ KL --- $\tau$} \\
\hspace{0.4em}$0.004$ & 86.12 & 66.95 & 43.17 & 53.70 & 30.44 & 36.52 & 7.08 & 46.28 \\
\hspace{0.4em}$0.04$ & 84.49 & 65.48 & 40.53 & 55.75 & 32.61 & 36.51 & 9.01 & 46.34 \\
\hspace{0.4em}$0.4\,^{\dagger}$ & 76.61 & 71.73 & 48.19 & 56.79 & 35.78 & 39.51 & 10.28 & 48.41 \\
\hspace{0.4em}$4$ & 82.05 & 70.05 & 46.71 & 55.19 & 34.31 & 38.74 & 9.93 & 48.14 \\
\bottomrule
\end{tabularx}
\label{tab:baseline_tuning}
\end{table*}

\clearpage

\subsection{Multi-Seed Results}
\label{app:multiseed}
We further evaluate robustness across three random seeds (42, 43, and 44) for both mathematical reasoning and code generation. 
Table~\ref{tab:multiseed} reports every run together with the mean and sample standard deviation (STD). 
On mathematical reasoning, \method{} achieves the highest average performance, outperforming DPO and Off-RL by 3.91\% and 4.21\%, respectively. 
On code generation, \method{} likewise achieves the best overall average and the strongest mean performance on MBPP+ and BCB.

\begin{table*}[!t]
\centering
\caption{\textbf{Three-seed results on mathematical reasoning and code generation.}
Method rows report mean $\pm$ sample STD over seeds 42, 43, and 44, with individual runs shown below.}
\small
\renewcommand{\arraystretch}{1.10}

\textbf{(a) Mathematical Reasoning on Qwen3-4B-Base}

\vspace{0.4em}
\setlength{\tabcolsep}{1pt}
\begin{tabularx}{\textwidth}{l *{8}{C}}
\toprule
\textbf{Method}
& \textbf{GSM8K}
& \textbf{MATH500}
& \textbf{MATH-P}
& \textbf{GaoKao}
& \textbf{Olym.}
& \textbf{AMC}
& \textbf{AIME}
& \textbf{Avg.} \\
\midrule
DPO
& \meanstd{84.10}{0.92}
& \meanstd{68.25}{1.49}
& \meanstd{43.41}{0.24}
& \meanstd{53.86}{1.19}
& \meanstd{32.33}{0.37}
& \meanstd{38.02}{1.58}
& \meanstd{9.14}{0.59}
& \meanstd{47.02}{0.82} \\
\hspace{0.5em}Seed 42
& 85.05 & 69.35 & 43.66 & 55.13 & 32.76 & 39.39 & 9.58 & 47.85 \\
\hspace{0.5em}Seed 43
& 84.04 & 66.55 & 43.37 & 52.76 & 32.07 & 36.29 & 8.47 & 46.22 \\
\hspace{0.5em}Seed 44
& 83.22 & 68.85 & 43.19 & 53.70 & 32.17 & 38.38 & 9.37 & 46.98 \\
\midrule
Off-RL
& \meanstd{87.05}{0.25}
& \meanstd{67.01}{0.28}
& \meanstd{41.94}{0.55}
& \meanstd{53.91}{1.03}
& \meanstd{31.68}{0.88}
& \meanstd{36.76}{0.16}
& \meanstd{8.68}{0.16}
& \meanstd{46.72}{0.41} \\
\hspace{0.5em}Seed 42
& 86.80 & 66.70 & 41.42 & 52.76 & 30.72 & 36.91 & 8.50 & 46.26 \\
\hspace{0.5em}Seed 43
& 87.05 & 67.07 & 41.91 & 54.16 & 32.43 & 36.60 & 8.72 & 46.85 \\
\hspace{0.5em}Seed 44
& 87.30 & 67.25 & 42.52 & 54.77 & 31.91 & 36.77 & 8.82 & 47.05 \\
\midrule
\rowcolor{rowhl}
\textbf{\method{}}
& \textbf{\meanstd{90.35}{1.23}}
& \textbf{\meanstd{73.30}{1.15}}
& \textbf{\meanstd{47.32}{1.28}}
& \textbf{\meanstd{58.98}{1.77}}
& \textbf{\meanstd{35.13}{0.73}}
& \textbf{\meanstd{41.49}{1.19}}
& \textbf{\meanstd{9.95}{0.52}}
& \textbf{\meanstd{50.93}{1.04}} \\
\hspace{0.5em}Seed 42
& 91.60 & 74.50 & 48.40 & 59.60 & 35.40 & 42.80 & 10.10 & 51.77 \\
\hspace{0.5em}Seed 43
& 90.30 & 73.17 & 47.65 & 60.36 & 35.69 & 41.18 & 10.38 & 51.25 \\
\hspace{0.5em}Seed 44
& 89.15 & 72.22 & 45.90 & 56.98 & 34.30 & 40.49 & 9.38 & 49.77 \\
\bottomrule
\end{tabularx}

\vspace{1.0em}

\textbf{(b) Code Generation on Qwen3-4B-Base}

\vspace{0.4em}
\setlength{\tabcolsep}{5pt}
\begin{tabular}{lcccc}
\toprule
\textbf{Method}
& \textbf{MBPP+}
& \textbf{BCB}
& \textbf{LCB}
& \textbf{Avg.} \\
\midrule
DPO
& \meanstd{60.13}{0.13}
& \meanstd{38.47}{0.17}
& \meanstd{22.02}{0.15}
& \meanstd{40.21}{0.10} \\
\hspace{0.5em}Seed 42 & 60.15 & 38.62 & 22.20 & 40.32 \\
\hspace{0.5em}Seed 43 & 60.25 & 38.29 & 21.94 & 40.16 \\
\hspace{0.5em}Seed 44 & 60.00 & 38.49 & 21.93 & 40.14 \\
\midrule
Off-RL
& \meanstd{60.61}{0.66}
& \meanstd{38.66}{0.27}
& \textbf{\meanstd{23.48}{0.07}}
& \meanstd{40.92}{0.23} \\
\hspace{0.5em}Seed 42 & 60.05 & 38.90 & 23.41 & 40.79 \\
\hspace{0.5em}Seed 43 & 61.34 & 38.72 & 23.50 & 41.19 \\
\hspace{0.5em}Seed 44 & 60.45 & 38.36 & 23.54 & 40.78 \\
\midrule
\rowcolor{rowhl}
\textbf{\method{}}
& \textbf{\meanstd{61.90}{0.37}}
& \textbf{\meanstd{39.63}{0.31}}
& \meanstd{23.18}{0.05}
& \textbf{\meanstd{41.57}{0.15}} \\
\hspace{0.5em}Seed 42 & 61.74 & 39.95 & 23.19 & 41.63 \\
\hspace{0.5em}Seed 43 & 62.33 & 39.59 & 23.13 & 41.68 \\
\hspace{0.5em}Seed 44 & 61.64 & 39.34 & 23.23 & 41.40 \\
\bottomrule
\end{tabular}

\label{tab:multiseed}
\end{table*}
\end{document}